\documentclass{article}
\usepackage{iclr2027_conference,times}

\usepackage[table]{xcolor}
\usepackage{colortbl}
\usepackage{graphicx}
\usepackage{booktabs}
\usepackage{amsmath,amssymb}
\usepackage{pifont}
\usepackage{fancyvrb}
\usepackage{url}
\usepackage{wrapfig}
\usepackage[inline]{enumitem}
\usepackage{caption}
\usepackage{hyperref}
\hypersetup{colorlinks,
  citecolor=[rgb]{0.10,0.50,0.15},
  linkcolor=[rgb]{0.75,0.20,0.15},
  urlcolor=[rgb]{0.80,0.15,0.55}}
\usepackage[capitalize]{cleveref}
\crefname{section}{Section}{Sections}
\crefname{appsec}{Appendix}{Appendices}

\newcommand{\method}{HN-CLIP}
\definecolor{deltagreen}{HTML}{1A8026}
\definecolor{deltared}{HTML}{FF6B6B}
\definecolor{venuegray}{HTML}{7F807F}
\definecolor{tipink}{HTML}{FADCEB}
\definecolor{itblue}{HTML}{CBDCEB}
\definecolor{bandgray}{HTML}{F2F3F8}
\definecolor{codegray}{HTML}{58585A}
\definecolor{codebg}{HTML}{FAFAFA}
\newcommand{\up}[1]{\textcolor{deltagreen}{$\uparrow$\,#1}}
\newcommand{\down}[1]{\textcolor{venuegray}{$\downarrow$\,#1}}
\newcommand{\vtag}[1]{{\scriptsize\textcolor{venuegray}{[#1]}}}
\newcommand{\tih}{\colorbox{tipink}{\textit{Text$\rightarrow$Image}}}
\newcommand{\ith}{\colorbox{itblue}{\textit{Image$\rightarrow$Text}}}
\newcommand{\cmark}{\ding{51}}
\newcommand{\xmark}{\textcolor{deltared}{\ding{55}}}
\newcommand{\lhn}{\mathcal{L}_{\mathrm{HN}}}
\newcommand{\lclip}{\mathcal{L}_{\mathrm{CLIP}}}
\newcommand{\ltok}{\mathcal{L}_{\mathrm{tok}}}

\title{Which Negatives Matter?\\
Ask Your Text Encoder: Adaptive Similarity\\
Margins for Dense-Caption Retrieval}

\author{
\textbf{Haoyue Liu}\textsuperscript{1}
\quad
\textbf{Ye Chen}\textsuperscript{2}
\quad
\textbf{Zhichao Wang}\textsuperscript{1}
\quad
\textbf{Xiaoying Tang}\textsuperscript{1,3,\ensuremath{\dagger}}
\\[0.6em]
\textsuperscript{1}
School of Science and Engineering,
The Chinese University of Hong Kong, Shenzhen 518172, China
\\
\textsuperscript{2}
XJTU-POLIMI Joint School, Xi'an Jiaotong University, Xi'an 710049, China
\\
\textsuperscript{3}
Shenzhen Future Network of Intelligence Institute (FNii-Shenzhen)
}

\iclrfinalcopy 

\begin{document}

\begin{SaveVerbatim}{hncode}
 def hard_neg_clip_loss(img_n, txt_n, scale, gamma=0.5):
   # img_n, txt_n: L2-normalized [B, D]
   B = img_n.shape[0]
   sim = img_n @ txt_n.t()                  # [B, B]
   txt_sim = (txt_n @ txt_n.t()).detach()
   boost = gamma * txt_sim
   boost.fill_diagonal_(0.)                 # positives
   logits = scale * (sim + boost)
   targets = torch.arange(B, device=sim.device)
   return (F.cross_entropy(logits, targets)
         + F.cross_entropy(logits.t(), targets)) / 2.
\end{SaveVerbatim}

\maketitle

\fancyhead{}

\begin{abstract}
Dense-caption retrieval has recently been improved by introducing
segmentation, edge maps, LLM-filtered captions, and cross-modal modules
into contrastive fine-tuning. However, these methods largely inherit the
same InfoNCE objective, whose optimization can prematurely saturate under
a strong pre-trained initialization: on dense captions, the loss falls
below $10^{-3}$ on $80\%$ of batches within the first epoch, while its
gradient becomes \emph{numerically zero} in $47\%$ of measurements. We
find that this behavior is closely related to the large number of
near-duplicate captions in dense-caption benchmarks, where a few highly
similar negatives remain unresolved after the easy majority has already
been separated. As a remedy, we introduce \method{}, which uses the text
encoder's own text--text geometry to construct per-negative adaptive
similarity margins. Specifically, a detached caption-similarity matrix is
added to the negative logits, assigning larger margins to more similar
captions without mining, synthesizing, or resampling negatives. The
resulting objective requires only one caption-similarity matrix and a
masked logit addition during training, with no auxiliary data, additional
parameters, offline preprocessing, or inference-time overhead. Extensive
experiments on four dense-caption retrieval benchmarks show that
\method{} improves over the strongest competitors by $+2.5$--$+4.0$
R@1 while training $2.4\times$ faster than GOAL and $5.4\times$ faster
than StructXLIP. Moreover, the proposed objective improves all six tested
fine-tuning frameworks on the in-domain benchmarks and reaches the
strongest full-data baseline with only $20\%$ of the training data.
\end{abstract}

\section{Introduction}
\label{sec:intro}

Contrastively pre-trained vision-language models
(VLMs)~\citep{radford2021learning,zhai2023sigmoid} have become a standard
backbone for image--text retrieval. While their pre-training corpora mainly
contain short web captions, recent benchmarks such as
DOCCI~\citep{onoe2024docci}, DCI~\citep{urbanek2024picture}, and
Urban-1K~\citep{zhang2024long} require models to distinguish long,
detail-rich descriptions of visually similar scenes. This
\emph{dense-caption retrieval} setting has motivated a growing line of
methods built on Long-CLIP~\citep{zhang2024long}. FineLIP introduces a
cross-modal module~\citep{asokan2025finelip}, GOAL incorporates local
matching over segmented regions~\citep{choi2025goal}, SmartCLIP reweights
salient tokens~\citep{xie2025smartclip}, and StructXLIP further introduces
structural cues from edge maps and LLM-derived lexical
supervision~\citep{ruan2026structxlip}. Despite their different designs,
these methods largely focus on enriching \emph{what} supervision is
provided while retaining the same global contrastive objective.

\begin{figure}[t]
  \centering
  \includegraphics[width=0.84\textwidth]{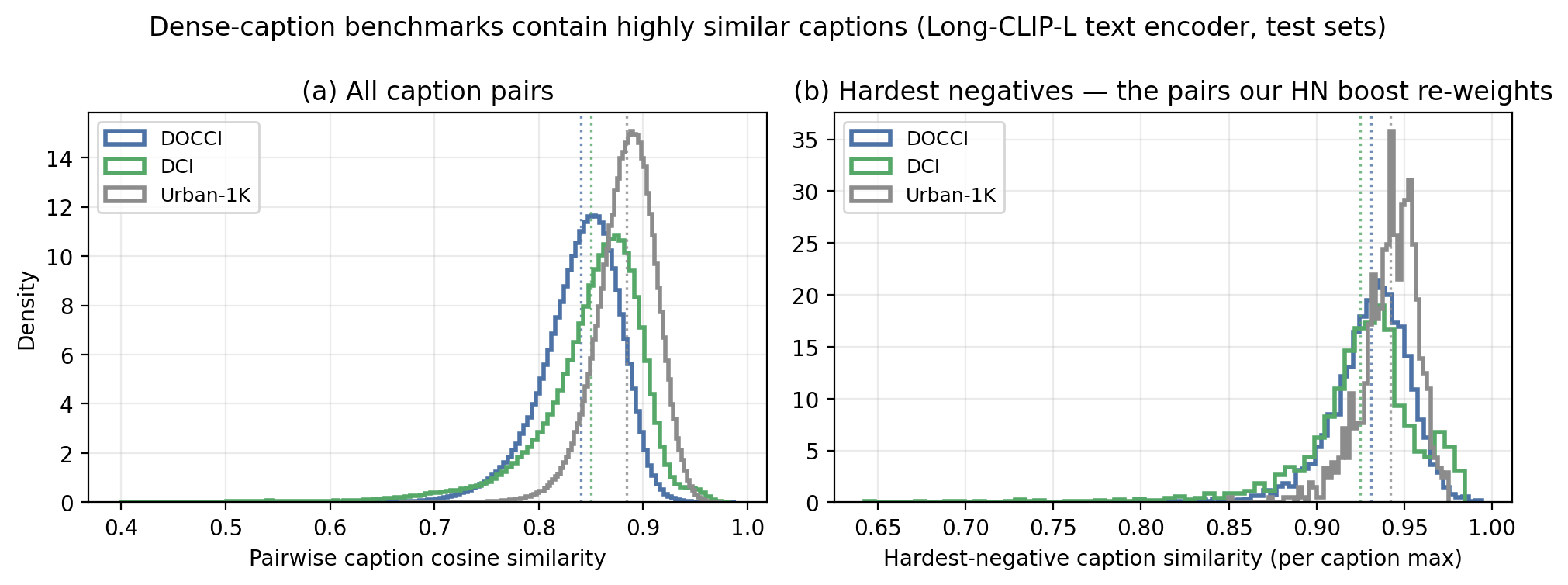}
  \caption{\textbf{Dense-caption benchmarks are dominated by hard
  negatives.} Distributions of (a) all pairwise caption-caption cosine
  similarities and (b) each caption's hardest-negative similarity,
  measured with the pre-trained Long-CLIP-L text encoder on the test sets.
  Dotted lines mark the means. The pairs in (b) receive the strongest
  boosts.}
  \label{fig:motivation}
\end{figure}

We instead investigate whether this objective remains effective during
dense-caption fine-tuning. Our analysis reveals a pronounced saturation
phenomenon: with a strong Long-CLIP initialization, the InfoNCE loss falls
below $10^{-3}$ on $80\%$ of measured batches within the first epoch, and
its gradient is \emph{numerically zero} in $47\%$ of measurements
(\cref{fig:gradient}). We trace this behavior to the caption geometry of
dense-caption benchmarks. Long descriptions are highly compositional and
often form near-duplicate pairs: under the pre-trained Long-CLIP text
encoder, each caption's hardest negative reaches $0.92$--$0.94$ cosine
similarity (\cref{fig:motivation}). A strong initialization therefore
separates the easy majority of negatives very quickly, causing the standard
objective to provide little gradient while the highly similar negatives
that determine R@1 remain unresolved. This observation raises a natural
question: \emph{can the contrastive objective itself adapt its training
pressure to the hard negatives already present in dense-caption batches?}

As a remedy, we introduce \method{}, a simple objective-level approach
that exploits the text encoder's own representation geometry to identify
hardness. Given a batch of image--caption pairs, \method{} computes the
caption--caption similarity matrix, detaches it from the computation graph,
masks its diagonal, and adds it to the negative logits with a single
coefficient $\gamma$. Each negative thus receives an adaptive margin
proportional to its caption similarity: easy negatives are nearly
unchanged, whereas near-duplicate captions must be separated by larger
margins before their loss vanishes. Unlike conventional hard-negative
strategies, \method{} does not mine, synthesize, or resample negatives;
the existing batch is left unchanged. Combined with a standard token-level
late-interaction term~\citep{yao2021filip}, the method introduces no
auxiliary inputs, preprocessing pipeline, architectural module, additional
parameters, or inference-time computation. Experiments on four
dense-caption benchmarks show that \method{} achieves the best R@1 in all
eight retrieval directions, outperforming the strongest competitors by
$+2.5$--$+4.0$ R@1. It also trains $2.4\times$ faster than GOAL and
$5.4\times$ faster than StructXLIP, surpasses the strongest full-data
baseline with only $20\%$ of the training data, and improves all six
tested fine-tuning frameworks on the in-domain benchmarks.

Our contributions are summarized as follows:
\begin{itemize}[leftmargin=*,itemsep=1pt,topsep=2pt]

\item \textbf{We identify a previously overlooked optimization issue in
dense-caption retrieval.}
Under a strong pre-trained initialization, InfoNCE rapidly separates the
easy majority of negatives and largely saturates while the highly similar
negatives that determine R@1 remain unresolved. We verify this behavior
through both caption-similarity statistics and direct gradient-dynamics
measurements.

\item \textbf{We introduce \method{}, a simple adaptive-margin objective
for dense-caption retrieval.}
\method{} converts the text encoder's own text--text geometry into
detached, per-negative similarity margins, assigning stronger training
pressure to more similar captions without negative mining, synthesis,
resampling, auxiliary data, architectural changes, or inference overhead.

\item \textbf{We provide extensive empirical evidence for the effectiveness
and generality of the proposed objective.}
Across four dense-caption benchmarks, \method{} achieves the best R@1 in
all eight retrieval directions with gains of $+2.5$--$+4.0$ over the
strongest competitors, trains substantially faster than machinery-based
methods, surpasses the strongest full-data baseline using only one fifth
of the training data, and improves all six tested fine-tuning frameworks
on the in-domain benchmarks. Gradient analysis further connects these
improvements to sustained optimization signal on hard negatives.

\end{itemize}

\section{Method}
\label{sec:method}

\begin{figure}[t]
  \centering
  \includegraphics[
    width=\textwidth,
    height=0.52\textwidth,
    keepaspectratio
  ]{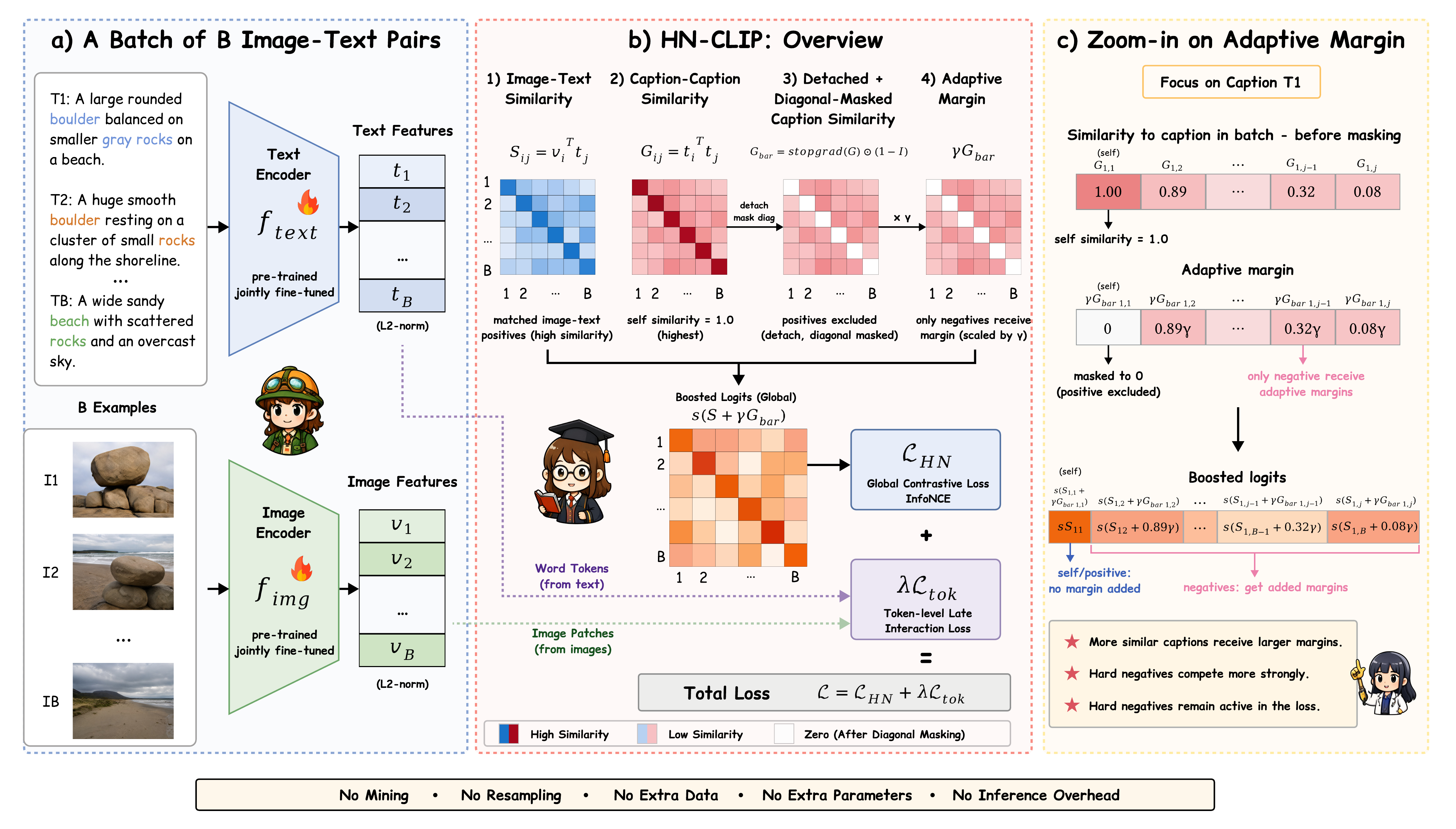}
  \caption{\textbf{Overview of \method{}.}
\textbf{(a)} A batch of image--text pairs is encoded by the jointly fine-tuned dual encoder.
\textbf{(b)} Caption--caption similarity provides detached, per-negative adaptive margins for the global contrastive loss, combined with the token-level term.
\textbf{(c)} More similar negatives receive larger margins, while the positive is unchanged.
\method{} requires no mining, resampling, extra data, additional parameters, or inference-time overhead.}
  \label{fig:overview}
\end{figure}

In this section, we introduce \method{}, which fine-tunes a pre-trained
dual encoder without auxiliary data views, preprocessing pipelines, or
architectural modules (\cref{fig:teaser,fig:overview}). Training combines a
text-similarity--boosted global objective with a standard token-level
late-interaction term:

\begin{wrapfigure}[24]{r}{0.44\textwidth}
  \centering
  \vspace{-18pt}  \includegraphics[width=\linewidth]{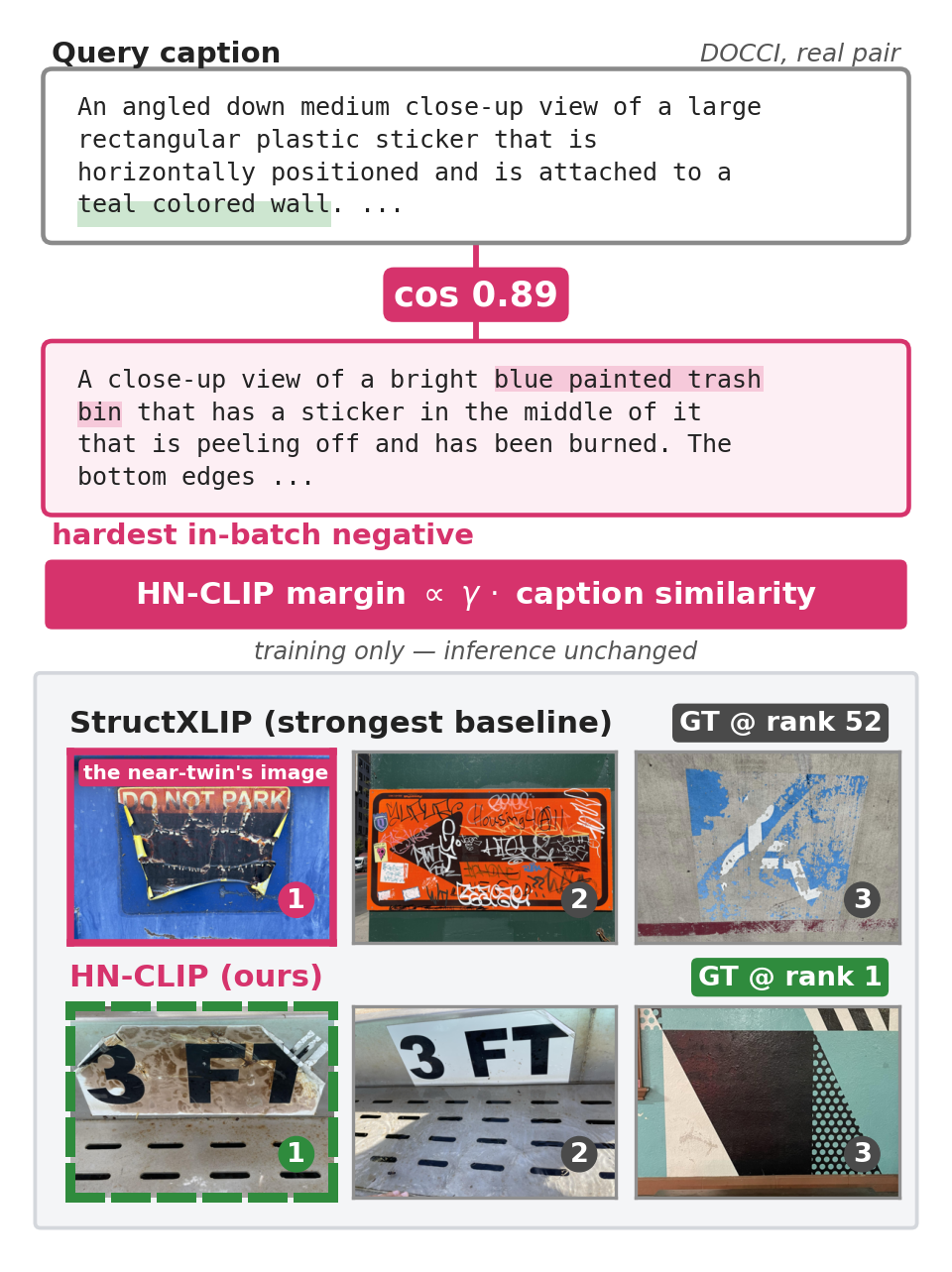}
  \caption{\textbf{Illustration of \method{}.} A real DOCCI query and its
  hardest in-batch negative (cos $0.89$). The strongest baseline ranks the ground truth 52nd (its top-1 (pink) is the near-twin's own
  image); \method{} ranks it first.}
  \label{fig:teaser}
  \vspace{-12pt}
\end{wrapfigure}

\begin{enumerate}[label=(\arabic*),leftmargin=*,itemsep=1pt,topsep=2pt]
\item The \emph{text-similarity--boosted objective}
  (\cref{sec:method-hn}), which converts the batch's own
  caption-similarity matrix into per-negative adaptive margins, keeping
  the loss and its gradient alive exactly on the hard negatives that
  decide retrieval.
\item A \emph{token-level alignment branch} (\cref{sec:method-tok}) that
  the boost makes usable, contributing an additional gain where captions
  are longest.
\end{enumerate}

We first quantify why the standard objective fails on dense captions
(\cref{sec:method-motivation}), then present both components, and close
with an empirical gradient-dynamics analysis (\cref{sec:method-analysis}).

\subsection{Preliminaries and motivation}
\label{sec:method-motivation}

Let $f_{\mathrm{img}}$ and $f_{\mathrm{txt}}$ map images and captions into a
shared $d$-dimensional space, and let
$\{(I_i, T_i)\}_{i=1}^{B}$ denote a batch of $B$ image-caption pairs. Let
$\mathbf{v}_i = f_{\mathrm{img}}(I_i)/\lVert f_{\mathrm{img}}(I_i)\rVert$ and
$\mathbf{t}_i = f_{\mathrm{txt}}(T_i)/\lVert f_{\mathrm{txt}}(T_i)\rVert$
denote the normalized embeddings, $S_{ij} = \mathbf{v}_i^{\!\top}\mathbf{t}_j$
the image-text similarity matrix, and $s$ the learned inverse temperature.
Standard fine-tuning minimizes the symmetric InfoNCE
loss~\citep{oord2018representation,radford2021learning}
\begin{equation}
\lclip = \tfrac{1}{2}\big[\mathrm{CE}(sS, y) + \mathrm{CE}(sS^{\top}, y)\big],
\label{eq:clip}
\end{equation}
where $\mathrm{CE}(Z, y)$ denotes row-wise softmax cross-entropy with
target indices $y_i = i$. The gradient of \cref{eq:clip} w.r.t.\ the logits
is the classic softmax residual~\citep{wang2021understanding}: each negative $j$ is repelled with force
proportional to its posterior probability
$p_{ij} = [\mathrm{softmax}(s S_{i,:})]_j$. When a strong pre-trained
initialization such as Long-CLIP already assigns the positive a large margin
over the \emph{easy majority} of negatives, those posteriors $p_{ij}$ become
negligible and can numerically underflow together with
the loss.

Dense-caption benchmarks make this failure mode extreme. Encoding the test
captions of DOCCI, DCI, and Urban-1K with the released Long-CLIP-L text
encoder, the mean \emph{pairwise} caption similarity is $0.84$, $0.85$, and
$0.88$, and the mean over each caption's \emph{hardest} companion is $0.93$,
$0.92$, and $0.94$ (\cref{fig:motivation}). Long descriptions of natural
scenes are compositional near-duplicates; the retrieval task is decided by a
handful of nearly-identical candidates. Empirically, at an effective batch of
128, $\lclip$ of a Long-CLIP-L initialization collapses below $10^{-5}$
within the first tens of steps of fine-tuning (\cref{fig:gradient}a): the
batch is ``solved'' with respect to easy negatives long before the model
separates the hard ones.

\subsection{Text-similarity--boosted hard negatives}
\label{sec:method-hn}

The diagnosis suggests the remedy: the objective must know \emph{which}
negatives are hard, and for captions this information is available for free.
We compute the text-text similarity matrix
$G_{ij} = \mathbf{t}_i^{\!\top}\mathbf{t}_j$ from embeddings the batch
already contains, detach it from the computation graph, zero its diagonal,
and add it to the logits of \cref{eq:clip}:
\begin{equation}
\lhn = \tfrac{1}{2}\big[\mathrm{CE}\big(s(S + \gamma\bar G),\, y\big)
      + \mathrm{CE}\big(s(S + \gamma\bar G)^{\top},\, y\big)\big],
\label{eq:hn}
\end{equation}
where $\bar G = \mathrm{stopgrad}(G) \odot (\mathbf{1} - \mathbb{I})$,
with $\odot$ the element-wise product, $\mathbf{1}$ the all-ones matrix,
$\mathbb{I}$ the identity (masking the positives), and $\gamma$ the single
hyperparameter (default $0.5$).

\paragraph{Interpretation as an adaptive margin.} Because $\bar G$ is
detached, \cref{eq:hn} is exactly \cref{eq:clip} evaluated on shifted logits:
negative $j$ of query $i$ competes with an additive handicap
$\gamma G_{ij}$ in its favor. Equivalently, the positive must beat every
negative by a margin \emph{proportional to how similar that negative's
caption is to its own}, a per-pair adaptive generalization of the fixed
additive margins used in metric learning. Easy negatives
($G_{ij}$ small) are almost unaffected; near-duplicate captions
($G_{ij}\!\to\!1$) keep producing loss and gradient until the model
separates them by the full margin. The softmax residual now concentrates
exactly on the pairs identified in \cref{fig:motivation}b (a formal
derivation of this gradient concentration is given in
Appendix~\ref{app:method-details}).

\medskip\noindent\textbf{Remark 2.1} (\emph{Why text--text
similarity})\textbf{.} \emph{Hardness could also be estimated from the
image-text scores $S$ being optimized, but early in training these are
exactly the quantities that are wrong, and reweighting by them makes the
supervision a function of the error it should correct. The text--text
geometry of a pre-trained encoder is instead accurate before the first
gradient step, symmetric across both retrieval directions, and the natural
space in which the benchmark difficulty manifests (\cref{fig:motivation}).
The stop-gradient removes any differentiable path to $\bar G$, so no step
is taken toward reshaping the model's own margins; it does not freeze
$\bar G$ across training, since we recompute it from the current encoder.
That residual drift is deliberate: \cref{tab:gamma}c shows it acts as an
implicit annealing of the margin, and that freezing it costs up to $7.2$
R@1.}\medskip

\paragraph{Cost.} Training overhead is one $B{\times}B$ matrix product and
one masked addition per step; there is no auxiliary forward pass, no extra
encoder, no offline extraction. Inference is byte-identical to the
underlying backbone. The complete reference implementation is 11 lines
(Appendix~\ref{app:method-details}).

\subsection{Token-level alignment branch}
\label{sec:method-tok}

The boost operates on global embeddings, and composes with supervision at a
finer granularity. Following FILIP~\citep{yao2021filip}, we add a
late-interaction term $\ltok$ that aligns each word token with its most
similar image patch (and vice versa), averaged over tokens, with in-batch
negatives. The full objective is
\begin{equation}
\mathcal{L} = \lhn + \lambda\,\ltok, \qquad \lambda = 1.
\label{eq:total}
\end{equation}
The two terms are complementary in a way our diagnosis predicts: $\lhn$
decides \emph{which} pairs the objective spends gradient on, $\ltok$
decides \emph{at what granularity} it is applied, and the second question
becomes material only once the first is answered, which
\cref{sec:exp-ablation} confirms. Both terms act at training time only:
no parameters are added and inference stays byte-identical.

\subsection{Gradient-dynamics analysis}
\label{sec:method-analysis}

\begin{figure}[t]
  \centering
  \includegraphics[width=\textwidth]{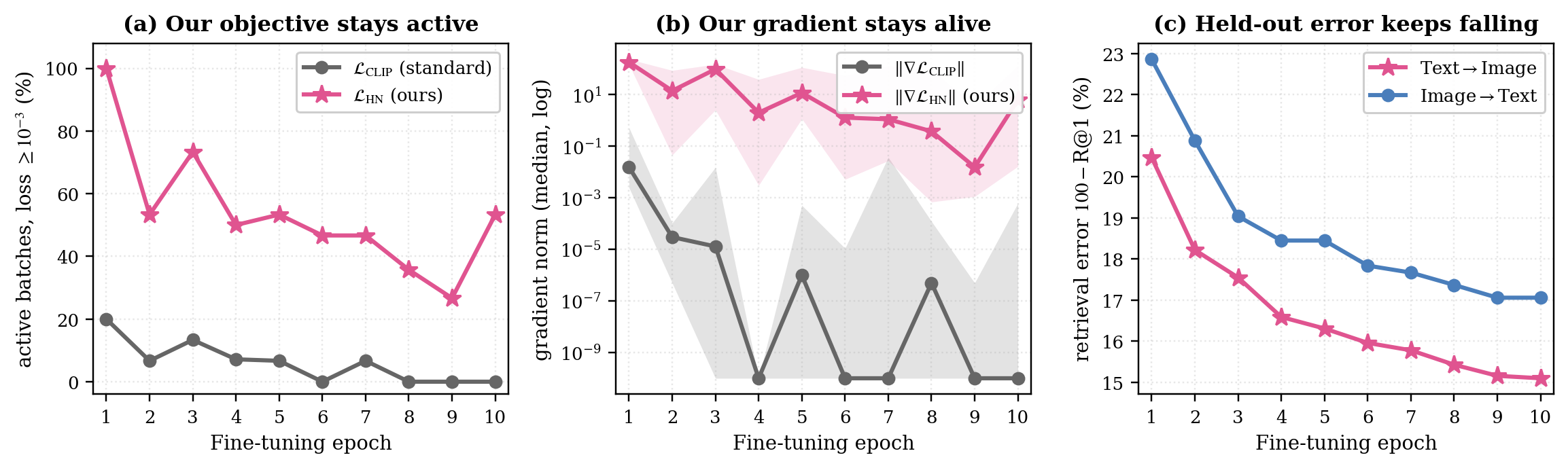}
  \caption{\textbf{Empirical gradient analysis} (Long-CLIP-L, DOCCI, 10
  epochs; losses and full-parameter gradients measured every 5 optimizer
  steps, 148 measurements). Standard InfoNCE declares $80\%$ of batches solved
  within the first epoch and its gradient is numerically zero in $47\%$ of
  measurements (median zero at five epochs; plotted clamped to $10^{-10}$).
  The boosted loss keeps at least $26\%$ of batches active in every epoch and,
  where the standard gradient is nonzero, exceeds it by
  $10^{2}$--$10^{6}\times$ in per-epoch median, with interquartile bands
  disjoint at nine epochs, and the retrieval error it buys keeps falling.
  Its gradient-norm distribution is bimodal across batches; see
  \cref{app:gradient}.}
  \label{fig:gradient}
\end{figure}

An objective that remains unsaturated supplies gradient diversity after the
main loss converges~\citep[cf.\ gradient starvation;][]{pezeshki2021gradient}, the
information-theoretic motivation behind
StructXLIP's auxiliary losses~\citep{ruan2026structxlip}. Our boost
realizes the same mechanism \emph{without any auxiliary view}, by reshaping
the main objective to reduce premature saturation and keep hard negatives
active longer. We verify this directly. During a 10-epoch fine-tuning run
of Long-CLIP-L on DOCCI (effective batch 128) we measure, every 5 optimizer
steps, both losses on the current batch, the norms of their gradients
$\lVert\nabla_{\theta}\lclip\rVert$ and $\lVert\nabla_{\theta}\lhn\rVert$
w.r.t.\ all model parameters $\theta$,
and the cosine similarity between the two gradients (\cref{fig:gradient}).

Three observations support the design. \textbf{(i) Saturation vs.\
persistence:} within the first epoch, $\lclip$ already falls below
$10^{-3}$ on $80\%$ of measured batches, and from epoch 6 on it does so on
essentially all of them: the standard objective simply runs out of work.
In contrast, the fraction of batches on which $\lhn$ still produces loss never falls
below $26\%$ (\cref{fig:gradient}a). The standard gradient is exactly zero
in $47\%$ of measurements; on the rest, $\lhn$ exceeds it by
$10^{2}$--$10^{6}\times$ in per-epoch median (\cref{fig:gradient}b).
\textbf{(ii) Compatibility:} the two gradients remain positively correlated
throughout ($\mathrm{cos}$ $\mu{=}0.47$, $\sigma{=}0.36$), i.e., the boost
steers optimization further along directions compatible with the original
objective rather than against it. \textbf{(iii) Utility:} the persistent
gradient is not noise: held-out retrieval error on DOCCI decreases
monotonically from $20.5\%$ to $15.1\%$ (T$\rightarrow$I) and $22.9\%$ to
$17.1\%$ (I$\rightarrow$T) across the same run (\cref{fig:gradient}c),
epochs after $\lclip$ has flattened. The standard objective largely stops
providing gradient signal;
the boosted one does not. Appendix~\ref{app:gradient} repeats this analysis at four
training-set scales on two datasets with identical conclusions.

\section{Experiments}
\label{sec:exp}

In this section, we carry out experiments to address the following
questions:

\noindent\begin{minipage}{\textwidth}
\begin{itemize}[leftmargin=*,itemsep=1pt,topsep=2pt]
\item \textbf{Q1}: Does \method{} outperform machinery-based fine-tuning
  methods on dense-caption retrieval? See \cref{sec:exp-main}.
\item \textbf{Q2}: Does the boost consistently improve existing
  fine-tuning frameworks in-domain? See \cref{sec:exp-plug}.
\item \textbf{Q3}: Do the sharpened decision boundaries transfer across
  domains and data scales? See \cref{sec:exp-transfer}.
\item \textbf{Q4}: Which component drives the gains, and how sensitive is
  the single hyperparameter $\gamma$? See \cref{sec:exp-ablation}.
\end{itemize}
\end{minipage}

\noindent Additional results (full-resolution numbers at deeper ranks (R@25/50),
seed replication, per-direction convergence, and training-efficiency
measurements) can be found in Appendices~\ref{app:impl}
and~\ref{app:analyses}.

\subsection{Experiment Setup}
\label{sec:exp-setup}

\begin{itemize}[leftmargin=*,itemsep=2pt,topsep=2pt]
\item \textbf{Benchmarks.} DOCCI~\citep{onoe2024docci} provides 15k images
  with highly discriminative human descriptions (123 words on average;
  9.5k train / 5.1k test). DCI~\citep{urbanek2024picture} contains 7.4k images
  with dense, mask-aligned captions (5.4k train / 2k test); following the
  protocol of \citet{choi2025goal} we additionally report \emph{Long-DCI},
  which evaluates the same models with the full-length captions.
  Urban-1K~\citep{zhang2024long} is a 1k-image test-only benchmark of
  urban scenes; as it provides no training split, models are fine-tuned on
  Visual Genome paragraph captions~\citep{krause2017hierarchical} and
  evaluated by transfer.
\item \textbf{Compared methods.} We compare against the released
  Long-CLIP~\citep{zhang2024long} and against
  FineLIP~\citep{asokan2025finelip}, GOAL~\citep{choi2025goal}, and
  StructXLIP~\citep{ruan2026structxlip}, all fine-tuned with their
  official code from the same Long-CLIP-L initialization with an identical
  budget.
\item \textbf{Implementation.} All experiments use the Long-CLIP-L
  backbone (ViT-L/14; text encoder stretched to 248 tokens). \method{} uses
  AdamW~\citep{loshchilov2017decoupled} (learning rate $2{\times}10^{-6}$,
  cosine schedule), effective
  batch 128, 10 epochs, $\gamma{=}0.5$, on Ascend 910B accelerators.
  Baselines use their official hyperparameters under the same backbone,
  batch size, and epoch budget; auxiliary inputs required by GOAL and
  StructXLIP are generated with their official pipelines. All methods
  are trained for the same 10-epoch budget under a shared evaluation
  protocol. The reported ranking is unchanged under last-epoch evaluation;
  a second seed on Long-DCI also preserves the R@1 ranking (mean
  $|\Delta|{=}0.09$ R@1 for \method{}). We report Recall@K ($K{=}1/5/10$) for
  T$\rightarrow$I and I$\rightarrow$T retrieval. Every \method{} number
  reported uses the full objective of \cref{eq:total} at
  $\gamma{=}0.5,\lambda{=}1$ on all four benchmarks, with no per-benchmark
  recipe and each term isolated in \cref{tab:gamma}b. Full details are
  in Appendix~\ref{app:impl}.
\end{itemize}

\begin{table}[t]
\centering
\caption{\textbf{Cross-modal retrieval performance of CLIP-based
fine-tuning methods on four dense-caption benchmarks.} We report Recall@K
(\%) on both \tih{} and \ith{} settings. All fine-tuned methods start from
the same Long-CLIP-L backbone with an identical training budget; Long-CLIP
denotes the released checkpoint. Best results in \textbf{bold}; second best
\underline{underlined}. $\Delta$ denotes the margin over the best
competitor per column, with gain in \textcolor{deltagreen}{$\uparrow$
green}.}
\label{tab:main}
\setlength{\tabcolsep}{2.6pt}
\scriptsize
\resizebox{\textwidth}{!}{%
\begin{tabular}{l|ccc|ccc|ccc|ccc}
\toprule
& \multicolumn{6}{c|}{\cellcolor{bandgray}\textbf{DOCCI}} & \multicolumn{6}{c}{\cellcolor{bandgray}\textbf{DCI}} \\
Method & \cellcolor{tipink}R@1 & \cellcolor{tipink}R@5 & \cellcolor{tipink}R@10 & \cellcolor{itblue}R@1 & \cellcolor{itblue}R@5 & \cellcolor{itblue}R@10 & \cellcolor{tipink}R@1 & \cellcolor{tipink}R@5 & \cellcolor{tipink}R@10 & \cellcolor{itblue}R@1 & \cellcolor{itblue}R@5 & \cellcolor{itblue}R@10 \\
\midrule
Long-CLIP\vtag{ECCV'24} & 78.78 & 95.24 & 98.02 & 66.75 & 91.92 & 96.31 & 67.83 & 83.19 & 87.69 & 64.13 & 84.84 & 89.74 \\
FineLIP\vtag{CVPR'25} & 77.51 & 96.02 & 98.41 & 69.90 & 93.43 & 97.45 & 72.69 & 87.14 & 90.65 & 65.48 & 86.84 & 91.00 \\
GOAL\vtag{CVPR'25} & 81.53 & 97.02 & 98.80 & 80.86 & 96.24 & 98.63 & \underline{77.29} & \underline{90.25} & 93.30 & \underline{74.84} & 89.94 & 93.25 \\
StructXLIP\vtag{CVPR'26} & \underline{84.73} & \underline{97.69} & \underline{99.00} & \underline{82.61} & \underline{97.08} & \underline{98.71} & 75.84 & 89.94 & \underline{93.65} & 74.49 & \underline{90.05} & \underline{93.40} \\
\textbf{\method{}} & \textbf{88.25} & \textbf{98.45} & \textbf{99.43} & \textbf{86.24} & \textbf{98.12} & \textbf{99.22} & \textbf{80.69} & \textbf{92.40} & \textbf{95.10} & \textbf{78.84} & \textbf{91.90} & \textbf{94.65} \\
\midrule
$\Delta$ & \up{3.52} & \up{0.76} & \up{0.43} & \up{3.63} & \up{1.04} & \up{0.51} & \up{3.40} & \up{2.15} & \up{1.45} & \up{4.00} & \up{1.85} & \up{1.25} \\
\bottomrule
\end{tabular}
%
}

\vspace{2pt}

\resizebox{\textwidth}{!}{%
\begin{tabular}{l|ccc|ccc|ccc|ccc}
\toprule
& \multicolumn{6}{c|}{\cellcolor{bandgray}\textbf{Long-DCI}} & \multicolumn{6}{c}{\cellcolor{bandgray}\textbf{Urban-1K}} \\
Method & \cellcolor{tipink}R@1 & \cellcolor{tipink}R@5 & \cellcolor{tipink}R@10 & \cellcolor{itblue}R@1 & \cellcolor{itblue}R@5 & \cellcolor{itblue}R@10 & \cellcolor{tipink}R@1 & \cellcolor{tipink}R@5 & \cellcolor{tipink}R@10 & \cellcolor{itblue}R@1 & \cellcolor{itblue}R@5 & \cellcolor{itblue}R@10 \\
\midrule
Long-CLIP\vtag{ECCV'24} & 54.61 & 72.80 & 78.33 & 47.35 & 73.04 & 80.10 & 86.10 & 96.50 & 98.10 & 82.40 & 96.70 & 98.30 \\
FineLIP\vtag{CVPR'25} & 59.24 & 77.86 & 83.19 & 49.52 & 75.08 & 82.39 & 81.50 & 94.50 & 97.60 & 77.90 & 95.00 & 98.00 \\
GOAL\vtag{CVPR'25} & 74.11 & 92.73 & 95.75 & \underline{73.29} & 92.18 & \underline{95.78} & 84.40 & 96.80 & 98.60 & \underline{88.20} & \underline{97.00} & \underline{98.60} \\
StructXLIP\vtag{CVPR'26} & \underline{75.31} & \underline{93.15} & \underline{95.92} & 72.56 & \underline{92.65} & 95.75 & \underline{86.80} & \underline{97.60} & \underline{98.90} & 87.70 & 96.90 & 98.30 \\
\textbf{\method{}} & \textbf{78.90} & \textbf{93.85} & \textbf{96.34} & \textbf{76.24} & \textbf{92.86} & \textbf{95.90} & \textbf{90.20} & \textbf{98.20} & \textbf{99.20} & \textbf{90.70} & \textbf{98.20} & \textbf{99.30} \\
\midrule
$\Delta$ & \up{3.59} & \up{0.70} & \up{0.42} & \up{2.95} & \up{0.21} & \up{0.12} & \up{3.40} & \up{0.60} & \up{0.30} & \up{2.50} & \up{1.20} & \up{0.70} \\
\bottomrule
\end{tabular}
%
}
\end{table}

\subsection{A1: \method{} Achieves Competitive Performance without Any
Machinery}
\label{sec:exp-main}

\paragraph{The performance of \method{} surpasses all machinery-based
baselines.} \cref{tab:main} reports the main comparison. As shown, we can
observe that:
\begin{enumerate}[leftmargin=*,itemsep=1pt,topsep=2pt]
\item \method{} achieves the outright best result in all 24 columns and
  achieves the best R@1 in \emph{all eight}
  directions, improving over the strongest
  competitor by $+3.52/+3.63$ R@1 on DOCCI, $+3.40/+4.00$ on DCI,
  $+3.59/+2.95$ on Long-DCI, and $+3.40/+2.50$ on Urban-1K.
\item The comparison is instructive about \emph{where} the gain comes
  from: GOAL consumes segmentation masks, StructXLIP edge maps and
  LLM-built lexicons, FineLIP a cross-modal module, yet a plain dual
  encoder that refuses to ignore hard negatives outperforms all of them.
\item Consistent with \cref{sec:method-analysis}, the largest margins
  appear at R@1, where near-duplicates decide the outcome; at deeper ranks
  all methods approach ceiling and margins compress, while \method{} remains
  best across all Long-DCI deep-rank columns.
\end{enumerate}

\begin{figure}[t]
  \centering
  \includegraphics[width=0.9\textwidth]{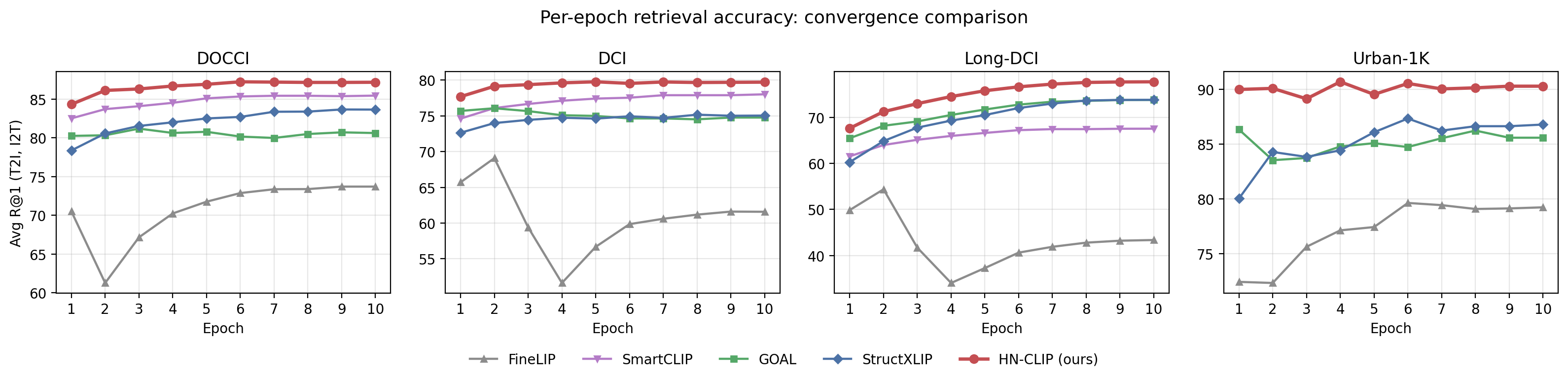}
  \caption{\textbf{Convergence comparison.} Average R@1 (T$\rightarrow$I,
  I$\rightarrow$T) per fine-tuning epoch. \method{}'s first epoch matches or
  exceeds most baselines' final accuracy on all four benchmarks.}
  \label{fig:convergence}
\end{figure}

\paragraph{\method{} accelerates the entire training trajectory.}
\cref{fig:convergence} plots per-epoch accuracy. On DOCCI the first epoch of
\method{} (84.3 average R@1) already exceeds the \emph{final} accuracy of
FineLIP, GOAL, and StructXLIP, and its second epoch (86.1) surpasses every
baseline. Hard-negative supervision does not merely raise the endpoint; it
accelerates the entire trajectory.

\begin{table}[t]
\centering
\caption{\textbf{Plug-and-play enhancement of our $\lhn$ on CLIP-based
fine-tuning.} Results on DOCCI and Long-DCI for \tih{} and \ith{}
retrieval. Upper: full-parameter fine-tuning; lower: parameter-efficient
tuning. Our method consistently boosts diverse CLIP variants, and the gains
grow with caption length. Best in \textbf{bold}, with gain in
\textcolor{deltagreen}{$\uparrow$ green}.}
\label{tab:plug}
\setlength{\tabcolsep}{2.2pt}
\scriptsize
\resizebox{\textwidth}{!}{%
\begin{tabular}{l|ccc|ccc|ccc|ccc}
\toprule
& \multicolumn{6}{c|}{\cellcolor{bandgray}\textbf{DOCCI}} & \multicolumn{6}{c}{\cellcolor{bandgray}\textbf{Long-DCI}} \\
Method & \cellcolor{tipink}R@1 & \cellcolor{tipink}R@5 & \cellcolor{tipink}R@10 & \cellcolor{itblue}R@1 & \cellcolor{itblue}R@5 & \cellcolor{itblue}R@10 & \cellcolor{tipink}R@1 & \cellcolor{tipink}R@5 & \cellcolor{tipink}R@10 & \cellcolor{itblue}R@1 & \cellcolor{itblue}R@5 & \cellcolor{itblue}R@10 \\
\midrule
Long-CLIP\vtag{ECCV'24} & 86.45 & 98.00 & 99.31 & 84.10 & 97.84 & 99.04 & 70.24 & 89.62 & 94.00 & 68.44 & 89.35 & 94.67 \\
+ our $\mathcal{L}_{\mathrm{HN}}$ & \textbf{88.31} & \textbf{98.75} & \textbf{99.45} & \textbf{86.39} & \textbf{98.08} & \textbf{99.20} & \textbf{78.43} & \textbf{92.52} & \textbf{95.27} & \textbf{76.02} & \textbf{91.78} & \textbf{94.76} \\
$\Delta$ & \up{1.86} & \up{0.75} & \up{0.14} & \up{2.29} & \up{0.24} & \up{0.16} & \up{8.19} & \up{2.90} & \up{1.27} & \up{7.58} & \up{2.43} & \up{0.09} \\
\addlinespace[1pt]
FineLIP\vtag{CVPR'25} & 77.51 & 96.02 & 98.41 & 69.90 & 93.43 & 97.45 & 59.24 & 77.86 & 83.19 & 49.52 & 75.08 & 82.39 \\
+ our $\mathcal{L}_{\mathrm{HN}}$ & \textbf{85.47} & \textbf{97.71} & \textbf{99.04} & \textbf{83.88} & \textbf{97.41} & \textbf{98.82} & \textbf{74.92} & \textbf{90.52} & \textbf{93.62} & \textbf{73.39} & \textbf{89.87} & \textbf{93.23} \\
$\Delta$ & \up{7.96} & \up{1.69} & \up{0.63} & \up{13.98} & \up{3.98} & \up{1.37} & \up{15.68} & \up{12.66} & \up{10.43} & \up{23.87} & \up{14.79} & \up{10.84} \\
\addlinespace[1pt]
GOAL\vtag{CVPR'25} & 81.53 & 97.02 & 98.80 & 80.86 & 96.24 & 98.63 & 74.11 & 92.73 & 95.75 & 73.29 & 92.18 & 95.78 \\
+ our $\mathcal{L}_{\mathrm{HN}}$ & \textbf{86.24} & \textbf{98.14} & \textbf{99.35} & \textbf{84.86} & \textbf{97.69} & \textbf{99.20} & \textbf{84.27} & \textbf{94.99} & \textbf{96.57} & \textbf{82.28} & \textbf{94.26} & \textbf{96.18} \\
$\Delta$ & \up{4.71} & \up{1.12} & \up{0.55} & \up{4.00} & \up{1.45} & \up{0.57} & \up{10.16} & \up{2.26} & \up{0.82} & \up{8.99} & \up{2.08} & \up{0.40} \\
\addlinespace[1pt]
StructXLIP\vtag{CVPR'26} & 84.73 & 97.69 & 99.00 & 82.61 & 97.08 & 98.71 & 75.31 & 93.15 & 95.92 & 72.56 & 92.65 & 95.75 \\
+ our $\mathcal{L}_{\mathrm{HN}}$ & \textbf{85.45} & \textbf{97.92} & \textbf{99.12} & \textbf{83.31} & \textbf{97.45} & \textbf{98.92} & \textbf{82.04} & \textbf{94.18} & \textbf{96.23} & \textbf{78.98} & \textbf{93.22} & \textbf{95.65} \\
$\Delta$ & \up{0.72} & \up{0.23} & \up{0.12} & \up{0.70} & \up{0.37} & \up{0.21} & \up{6.73} & \up{1.03} & \up{0.31} & \up{6.42} & \up{0.57} & \down{0.10} \\
\midrule
LoRA\vtag{ICLR'22} & 80.49 & 96.16 & 98.55 & 77.96 & 95.80 & 98.04 & 60.08 & 80.68 & 86.86 & 58.97 & 80.02 & 86.53 \\
+ our $\mathcal{L}_{\mathrm{HN}}$ & \textbf{82.82} & \textbf{97.14} & \textbf{98.86} & \textbf{80.84} & \textbf{96.49} & \textbf{98.63} & \textbf{64.87} & \textbf{83.09} & \textbf{88.35} & \textbf{62.79} & \textbf{81.77} & \textbf{87.31} \\
$\Delta$ & \up{2.33} & \up{0.98} & \up{0.31} & \up{2.88} & \up{0.69} & \up{0.59} & \up{4.79} & \up{2.41} & \up{1.49} & \up{3.82} & \up{1.75} & \up{0.78} \\
\addlinespace[1pt]
DoRA\vtag{ICML'24} & 80.76 & 96.25 & 98.61 & 78.20 & 95.90 & 98.10 & 60.76 & 81.19 & 87.48 & 59.65 & 80.31 & 87.13 \\
+ our $\mathcal{L}_{\mathrm{HN}}$ & \textbf{83.41} & \textbf{97.33} & \textbf{99.00} & \textbf{81.41} & \textbf{96.65} & \textbf{98.67} & \textbf{65.46} & \textbf{83.57} & \textbf{88.61} & \textbf{63.38} & \textbf{82.15} & \textbf{87.56} \\
$\Delta$ & \up{2.65} & \up{1.08} & \up{0.39} & \up{3.21} & \up{0.75} & \up{0.57} & \up{4.70} & \up{2.38} & \up{1.13} & \up{3.73} & \up{1.84} & \up{0.43} \\
\bottomrule
\end{tabular}
%
}
\end{table}

\subsection{A2: The Boost Consistently Improves Existing Frameworks In-Domain}
\label{sec:exp-plug}

\paragraph{Consistent in-domain plug-and-play improvement.}
Because $\lhn$ only modifies the logit matrix of the global contrastive
term, it can replace that term inside any fine-tuning framework. Following
the protocol of \citet{ruan2026structxlip}, we integrate it into the official
training code of Long-CLIP, FineLIP, GOAL, and StructXLIP, and into
LoRA~\citep{hu2021lora} and DoRA~\citep{liu2024dora} adapters on the same
backbone. \cref{tab:plug} shows consistent in-domain gains: every tested
framework improves on both benchmarks, and R@1 gains scale with caption length, from
$+0.70$--$+13.98$ on DOCCI up to $+3.73$--$+23.87$ on Long-DCI, exactly the
regime where hard negatives are most extreme
(cf.\ \cref{fig:motivation}). StructXLIP, whose own auxiliary losses
already target alignment quality, still benefits substantially; the
benefit extends to PEFT ($+3.7$--$+4.8$ R@1 for LoRA/DoRA on Long-DCI).
DCI results and the domain-transfer setting are analyzed in
Appendix~\ref{app:analyses}.

\subsection{A3: Sharpened Boundaries Transfer Across Domains and Scales}
\label{sec:exp-transfer}

\paragraph{Across domains.}
Hard-negative margins could in principle overfit dataset-specific caption
statistics; \cref{tab:cross} shows the opposite. Trained on DCI and
transferred to DOCCI, \method{} scores $85.00/83.14$ R@1, higher than
every baseline's \emph{in-domain} DOCCI result, and it leads every column
of the reverse and of the harder DOCCI$\rightarrow$Long-DCI transfer
(\cref{tab:cross,tab:s-cross}).

\begin{wrapfigure}[20]{r}{0.36\textwidth}
  \centering
  \vspace{-12pt}
  \includegraphics[width=\linewidth]{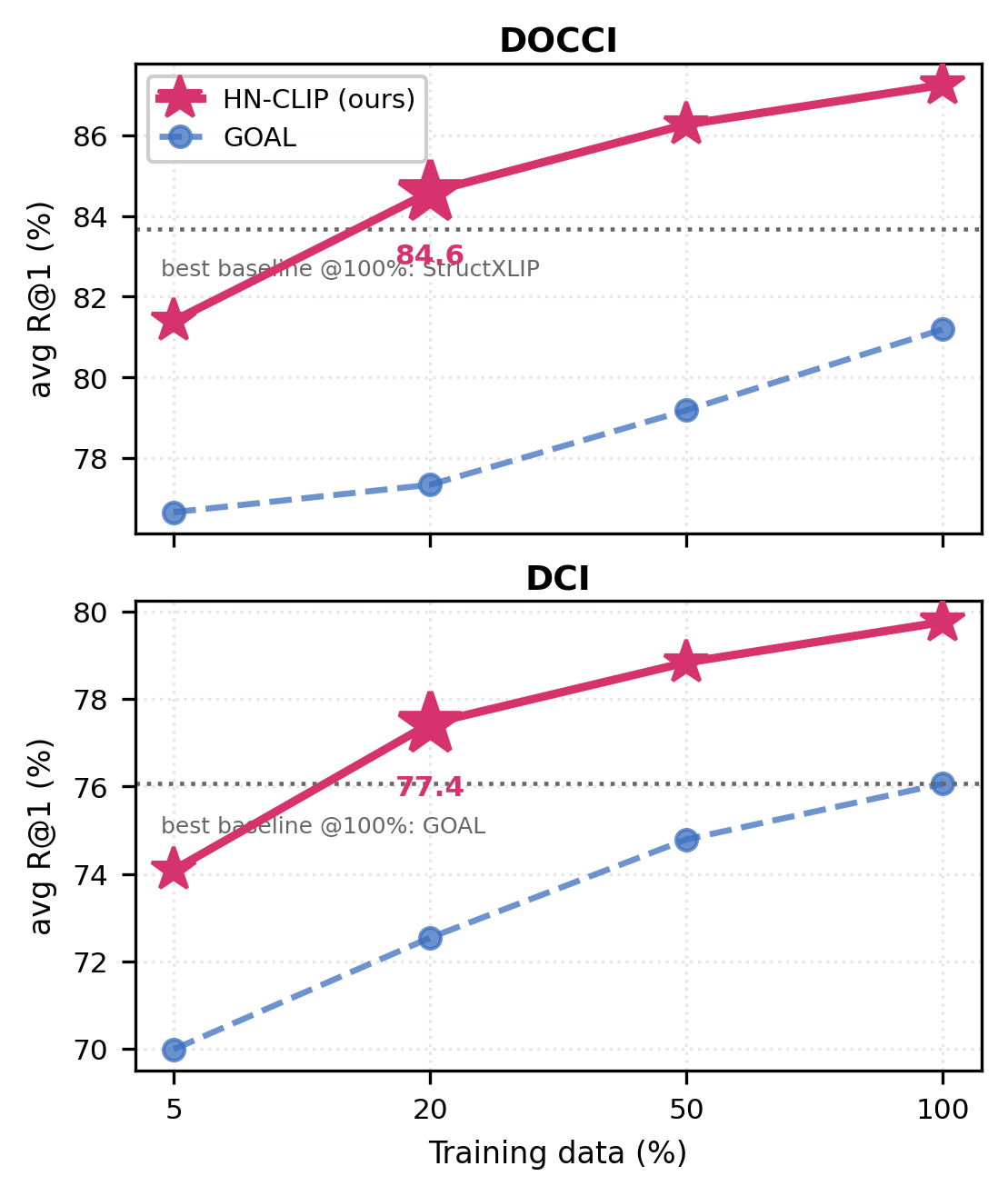}
  \caption{\textbf{Sample efficiency} (avg R@1, identical subsets).
  \method{} at $20\%$ of the data already clears the strongest baseline
  trained on $100\%$ (dotted line, from \cref{tab:main}); full numbers in
  \cref{tab:s-sampeff}.}
  \label{fig:sampeff-main}
  \vspace{-10pt}
\end{wrapfigure}

\paragraph{Across data scales.}
\cref{fig:sampeff-main} varies the training fraction on identical
subsets. \method{} leads at every fraction, the margin \emph{widens} as
data shrinks to $20\%$ ($+7.3$ R@1), and a fifth of the data already beats
GOAL trained on all of it. When gradient steps are scarce, spending them
on informative negatives matters most.

\subsection{A4: Ablation Studies}
\label{sec:exp-ablation}

\begin{table}[t]
\centering
\caption{\textbf{Ablations.} \textbf{(a)}~Sensitivity to the boost
strength $\gamma$ (Recall@1): every $\gamma\in[0.25,1]$ beats
$\gamma{=}0$ in-domain, while under transfer (Urban-1K) milder boosts
generalize better. \textbf{(b)}~Loss components under an identical recipe;
only the loss changes. \textbf{(c)}~$\bar G$ recomputed from the live
encoder (default) vs.\ frozen at initialization; R@1/5/10 in
\cref{tab:frozeng}. Best per column in \textbf{bold}.}
\label{tab:gamma}
{\footnotesize (a) boost strength $\gamma$\hfill\phantom{.}}
\vspace{1pt}

\setlength{\tabcolsep}{10pt}
\scriptsize
\resizebox{\textwidth}{!}{%
\begin{tabular}{l|cc|cc|cc|cc|c}
\toprule
& \multicolumn{2}{c|}{\cellcolor{bandgray}\textbf{DOCCI}} & \multicolumn{2}{c|}{\cellcolor{bandgray}\textbf{DCI}} & \multicolumn{2}{c|}{\cellcolor{bandgray}\textbf{Long-DCI}} & \multicolumn{2}{c|}{\cellcolor{bandgray}\textbf{Urban-1K}} & \cellcolor{bandgray} \\
$\gamma$ & \cellcolor{tipink}R@1 & \cellcolor{itblue}R@1 & \cellcolor{tipink}R@1 & \cellcolor{itblue}R@1 & \cellcolor{tipink}R@1 & \cellcolor{itblue}R@1 & \cellcolor{tipink}R@1 & \cellcolor{itblue}R@1 & \cellcolor{bandgray}\textbf{Avg} \\
\midrule
$\gamma{=}0$ & 86.45 & 84.29 & 79.24 & 77.54 & 70.96 & 68.08 & \textbf{92.00} & 92.90 & \cellcolor{bandgray}81.43 \\
$\gamma{=}0.25$ & 87.82 & \textbf{86.63} & \textbf{80.69} & \textbf{79.19} & 76.37 & 74.69 & 91.70 & \textbf{93.10} & \cellcolor{bandgray}\textbf{83.77} \\
$\gamma{=}0.5$ (default) & \textbf{88.25} & 86.24 & \textbf{80.69} & 78.84 & 78.90 & 76.24 & 90.20 & 90.70 & \cellcolor{bandgray}83.76 \\
$\gamma{=}0.75$ & 88.20 & 85.73 & 80.64 & 78.39 & 79.82 & \textbf{76.59} & 88.80 & 89.70 & \cellcolor{bandgray}83.48 \\
$\gamma{=}1.0$ & 87.98 & 85.02 & 80.04 & 78.29 & \textbf{80.05} & 76.48 & 88.20 & 88.30 & \cellcolor{bandgray}83.05 \\
\bottomrule
\end{tabular}
%
}

\vspace{6pt}
{\footnotesize (b) loss components\hfill\phantom{.}}
\vspace{1pt}

\setlength{\tabcolsep}{6pt}
\renewcommand{\arraystretch}{1.12}
\scriptsize
\resizebox{\textwidth}{!}{%
\begin{tabular}{cc|ccc|ccc|ccc|ccc|c}
\toprule
& & \multicolumn{6}{c|}{\cellcolor{bandgray}\textbf{DCI}} & \multicolumn{6}{c|}{\cellcolor{bandgray}\textbf{Long-DCI}} & \cellcolor{bandgray} \\
$\ltok$ & $\lhn$ & \cellcolor{tipink}R@1 & \cellcolor{tipink}R@5 & \cellcolor{tipink}R@10 & \cellcolor{itblue}R@1 & \cellcolor{itblue}R@5 & \cellcolor{itblue}R@10 & \cellcolor{tipink}R@1 & \cellcolor{tipink}R@5 & \cellcolor{tipink}R@10 & \cellcolor{itblue}R@1 & \cellcolor{itblue}R@5 & \cellcolor{itblue}R@10 & \cellcolor{bandgray}\textbf{Avg} \\
\midrule
\xmark & \xmark & 79.39 & 91.60 & 94.65 & 78.34 & 92.85 & 95.25 & 70.24 & 89.62 & 94.00 & 68.44 & 89.35 & 94.67 & \cellcolor{bandgray}74.10 \\
\cmark & \xmark & 79.24 & 91.55 & 94.25 & 77.54 & \textbf{92.95} & \textbf{95.35} & 70.96 & 89.52 & 93.95 & 68.08 & 89.24 & 94.25 & \cellcolor{bandgray}73.95 \\
\xmark & \cmark & \textbf{81.04} & 92.20 & 95.00 & 78.54 & 91.95 & 94.60 & 78.43 & 92.52 & 95.27 & 76.02 & 91.78 & 94.76 & \cellcolor{bandgray}78.51 \\
\cmark & \cmark & 80.69 & \textbf{92.40} & \textbf{95.10} & \textbf{78.84} & 91.90 & 94.65 & \textbf{78.90} & \textbf{93.85} & \textbf{96.34} & \textbf{76.24} & \textbf{92.86} & \textbf{95.90} & \cellcolor{bandgray}\textbf{78.67} \\
\bottomrule
\end{tabular}
%
}

\vspace{6pt}
{\footnotesize (c) $\bar G$ recomputed vs.\ frozen\hfill\phantom{.}}
\vspace{1pt}

\setlength{\tabcolsep}{10pt}
\renewcommand{\arraystretch}{1.0}
\scriptsize
\resizebox{\textwidth}{!}{%
\begin{tabular}{l|cc|cc|cc|cc|c}
\toprule
& \multicolumn{2}{c|}{\cellcolor{bandgray}\textbf{DOCCI}} & \multicolumn{2}{c|}{\cellcolor{bandgray}\textbf{DCI}} & \multicolumn{2}{c|}{\cellcolor{bandgray}\textbf{Long-DCI}} & \multicolumn{2}{c|}{\cellcolor{bandgray}\textbf{Urban-1K}} & \cellcolor{bandgray} \\
$\bar G$ source & \cellcolor{tipink}R@1 & \cellcolor{itblue}R@1 & \cellcolor{tipink}R@1 & \cellcolor{itblue}R@1 & \cellcolor{tipink}R@1 & \cellcolor{itblue}R@1 & \cellcolor{tipink}R@1 & \cellcolor{itblue}R@1 & \cellcolor{bandgray}\textbf{Avg} \\
\midrule
Recomputed $\bar G$ (default) & \textbf{88.25} & \textbf{86.24} & \textbf{80.69} & \textbf{78.84} & \textbf{78.90} & \textbf{76.24} & \textbf{90.20} & \textbf{90.70} & \textbf{83.76} \\
Frozen $\bar G$ (at init) & 85.86 & 82.43 & 77.79 & 75.84 & 77.93 & 74.75 & 87.30 & 83.50 & 80.68 \\
\midrule
$\Delta$ (default $-$ frozen) & \up{2.39} & \up{3.81} & \up{2.90} & \up{3.00} & \up{0.97} & \up{1.49} & \up{2.90} & \up{7.20} & \up{3.08} \\
\bottomrule
\end{tabular}
%
}

\end{table}

\paragraph{Loss components.}
\cref{tab:gamma}b isolates each term under a strictly identical recipe,
and the pattern directly tests our diagnosis. Token-level supervision
\emph{on its own} leaves accuracy where plain fine-tuning left it, exactly
what \cref{sec:method-analysis} predicts, since a finer granularity
inherits the same uniform treatment of negatives and saturates with it.
Granularity is not the bottleneck; negative weighting is. Once $\lhn$
removes that bottleneck both counts change: alone it produces the bulk of
the improvement ($+1.7/+0.2$ R@1 on DCI, $+8.2/+7.6$ on Long-DCI), and the
inert token branch now contributes $+0.47/+0.22$ on the hardest benchmark.
\emph{The boost is what makes finer supervision usable.}

\paragraph{Strength of the boost.}
The sweep over our only hyperparameter (\cref{tab:gamma}a) is
interpretable rather than tuned. Every in-domain
$\gamma\in\{0.25,\dots,1.0\}$ beats $\gamma{=}0$ by up to $+9.1$ R@1,
while the spread \emph{inside} the plateau is only $0.7$ average R@1,
far smaller than the gain from making $\gamma$ nonzero, so the choice of
$\gamma$ matters far less than making it nonzero. The endpoints are
informative: the hardest evaluation (Long-DCI) peaks at the strongest
boosts ($\gamma{=}0.75$--$1.0$), the Urban-1K transfer at the mildest ($\gamma\!\leq\!0.25$), exactly what a mechanism sharpening boundaries on the
\emph{training} distribution predicts. Our untuned default $\gamma{=}0.5$
lands within $0.03$ R@1 of the best setting under either aggregation---it
trails $\gamma{=}0.25$ by $0.01$ on the four-benchmark average and
$\gamma{=}0.75$ by $0.03$ on the three in-domain benchmarks---sits within
$0.7$ average R@1 of every in-domain benchmark optimum, and since $s$ and $\gamma$ act only through their
product (Appendix~\ref{app:deriv}), the plateau covers a corresponding range of
temperatures.

\paragraph{The drift of $\bar G$ is load-bearing.}
Because the text encoder trains, the recomputed $\bar G$ drifts across
steps; \cref{tab:gamma}c rebuilds the boost from a frozen copy of the
initial encoder to test whether that drift helps. The default wins the
average R@1 on all four benchmarks and all 24 columns. The frozen variant
peaks at epoch~1 on DCI and Urban-1K before declining, the signature of a
margin that never relaxes. Recomputation therefore acts as
an \emph{implicit annealing} of the boost, and an independent knob
corroborates the $\gamma$ sweep: the never-decaying margin loses most
exactly where milder boosts win, on transfer (Appendix~\ref{app:deriv}).

\section{Related Work}
\label{sec:related}

\paragraph{Long-text vision-language alignment.}
Contrastive image-text pre-training~\citep{radford2021learning,jia2021scaling,zhai2023sigmoid}
established the dual-encoder paradigm, but its $\sim$77-token window fits
paragraphs poorly. Long-CLIP~\citep{zhang2024long} relaxes the
constraint and is the standard backbone here; parallel work stretches the
window itself~\citep{najdenkoska2024tulip,wu2024lotlip} or re-captions the
corpus~\citep{fan2023improving,chen2024sharegpt4v}. On this backbone,
FineLIP~\citep{asokan2025finelip} inserts a cross-modal module,
GOAL~\citep{choi2025goal} adds local matching over segmented regions,
SmartCLIP~\citep{xie2025smartclip} re-weights salient tokens,
DreamLIP~\citep{zheng2024dreamlip} decomposes captions,
StructXLIP~\citep{ruan2026structxlip} aligns edge maps with
lexicon-filtered captions, FILIP~\citep{yao2021filip} (after
ColBERT~\citep{khattab2020colbert}) aligns word and patch tokens, and PEFT
adapters~\citep{hu2021lora,liu2024dora} update fewer weights. All enrich
\emph{what} is aligned while inheriting the InfoNCE objective unchanged;
\method{} instead repairs \emph{how} negatives inside that objective are
weighted, and so composes with diverse existing fine-tuning frameworks
(\cref{sec:exp-plug}).

\paragraph{Hard negatives in contrastive learning.}
The role of negatives in contrastive learning is well
established~\citep{oord2018representation,he2020momentum,chen2020simple,khosla2020supervised,wang2020understanding},
with prior work exploring importance-based
reweighting~\citep{robinson2020contrastive}, synthetic
mixing~\citep{kalantidis2020hard}, debiased
objectives~\citep{chuang2020debiased}, mined captions for
compositionality~\citep{yuksekgonul2022and,thrush2022winoground,hsieh2023sugarcrepe},
false-negative relabeling~\citep{li2023your,byun2024mafa}, hardest-negative
mining~\citep{faghri2017vse++}, and score-based
up-weighting~\citep{radenovic2023filtering}. These methods largely ask
\emph{which negatives to use}; \method{} instead asks \emph{where hardness
lives}, using the encoder's pre-trained text--text geometry
(\cref{sec:method-hn}). Unlike margin-based metric
learning~\citep{schroff2015facenet,wang2018cosface,deng2019arcface},
including embedding-dependent margins~\citep{kim2022adaface}, our margin is
per-pair, derived from text--text geometry, and detached. Unlike score-based reweighting such as
DiHT, which relies on the cross-modal scores being optimized, \method{} uses
text--text geometry available before the first update. To our knowledge, no
prior work converts caption-to-caption similarity into adaptive per-negative
margins or identifies the structural saturation of the standard objective in
dense-caption retrieval.

\section{Conclusion}
\label{sec:conclusion}

Dense-caption retrieval did not need more machinery; it needed an
objective that kept learning. We showed that InfoNCE largely saturates
within the first epoch, traced this failure to near-duplicate captions
and the absence of explicit caption-level hardness, and repaired it using
the same text--text geometry that exposed the problem. The resulting
training-time modification sets a new state of the art on four benchmarks,
matches or exceeds most baselines' final accuracy after a single epoch,
requires only one fifth of the training data to surpass the strongest
full-data baseline, and improves all six tested fine-tuning frameworks
on the in-domain benchmarks.

\subsubsection*{Reproducibility Statement}
The hard-negative objective $\lhn$ is implemented by the 11-line loss in
Appendix~\ref{app:method-details}; the token-level branch follows
FILIP~\citep{yao2021filip} and is specified in \cref{sec:method-tok}. All
remaining training components follow standard Long-CLIP fine-tuning. All hyperparameters, learning rates for every baseline, and
the plug-and-play integration points are specified in \cref{sec:exp} and
Appendix~\ref{app:impl}. All methods use the same training budget and
evaluation protocol. We further verify that the reported ranking is
unchanged under last-epoch evaluation (\cref{sec:exp}), and that the
Long-DCI R@1 ranking is preserved under a second random seed
(Appendix~\ref{app:analyses}). Code and training scripts will
be released upon publication.

\subsubsection*{AI Use Statement}
We used a large language model to polish our writing and to help write the
scripts that plot our figures and format our tables. The research question,
the method, the experiments, and all reported numbers are the authors' own;
we reviewed and verified all AI-assisted output and take responsibility for
the final content of this work.

\bibliography{main}
\bibliographystyle{iclr2027_conference}

\clearpage
\appendix
\setcounter{table}{0}
\setcounter{figure}{0}
\renewcommand{\thetable}{A\arabic{table}}
\renewcommand{\thefigure}{A\arabic{figure}}

\section*{Contents of Appendix}
\vspace{-4pt}
{\parindent=0pt
\newcommand{\apxdots}{\nobreak\leaders\hbox to 0.62em{\hss.\hss}\hfill\nobreak}
\newcommand{\apxpg}[1]{{\hypersetup{linkcolor=black}\pageref{#1}}}
\newcommand{\apxsec}[3]{\vspace{3.5pt}\textbf{#1}\hspace{1.1em}\textbf{#2}%
  \nobreak\hfill\nobreak\textbf{\apxpg{#3}}\par\vspace{0.5pt}}
\newcommand{\apxsub}[3]{\hspace{1.7em}#1\hspace{0.9em}#2\ \apxdots\ \apxpg{#3}\par}

\apxsec{A}{Dataset Details}{app:data}
\apxsub{A.1}{Benchmarks and splits}{app:data-splits}
\apxsub{A.2}{One batch, four benchmarks}{app:data-batch}
\apxsub{A.3}{What near-duplicates look like}{app:data-neardup}
\apxsub{A.4}{Statistics of the caption geometry}{app:data-stats}
\apxsec{B}{Method Details}{app:method-details}
\apxsub{B.1}{Reference implementation}{app:refimpl}
\apxsub{B.2}{Gradient concentration}{app:deriv}
\apxsec{C}{Extended Gradient Analysis}{app:gradient}
\apxsec{D}{Implementation Details and Computational Analysis}{app:impl}
\apxsub{D.1}{Training setup and backbone choice}{app:impl-setup}
\apxsub{D.2}{Plug-and-play integration}{app:impl-plug}
\apxsub{D.3}{Evaluation protocol}{app:impl-eval}
\apxsec{E}{Additional Experimental Analyses}{app:analyses}
\apxsub{E.1}{Plug-and-play, DOCCI and Long-DCI}{app:plug}
\apxsub{E.2}{Full-resolution ablations}{app:ablfull}
\apxsub{E.3}{Cross-domain evaluation}{app:cross}
\apxsub{E.4}{Robustness to the random seed}{app:seed}
\apxsub{E.5}{Sample efficiency, full numbers}{app:sampeff}
\apxsub{E.6}{Per-direction convergence}{app:conv}
\apxsub{E.7}{Results at deeper ranks}{app:deepranks}
\apxsec{F}{Additional Qualitative Results}{app:qual}
\par}
\vspace{6pt}

\section{Dataset Details}
\label{app:data}

\subsection{Benchmarks and splits}
\label{app:data-splits}
\cref{tab:s1} summarizes the four benchmarks, all members of the recent
family of hyper-detailed description
datasets~\citep{garg2024imageinwords}. All models are fine-tuned on
the official training split of each benchmark; Long-DCI evaluates the
DCI-trained models with full-length captions following the protocol of
GOAL, and Urban-1K is test-only, so models are fine-tuned on Visual Genome
paragraph captions and evaluated by transfer. The last two columns repeat
the measurement from \cref{sec:method-motivation}: with the pre-trained
Long-CLIP-L text encoder, the mean pairwise caption cosine similarity and
the mean similarity of each caption's \emph{hardest} companion, the
statistic that motivates and, unchanged, powers~$\lhn$.

\begin{table}[h]
\centering
\caption{\textbf{Benchmark statistics.} Mean caption length is measured on
the evaluation split in words; similarities use the pre-trained
Long-CLIP-L text encoder.}
\label{tab:s1}
\setlength{\tabcolsep}{3pt}
\renewcommand{\arraystretch}{1.15}
\scriptsize
\resizebox{0.75\textwidth}{!}{%
\begin{tabular}{l|cc|c|cc}
\toprule
Benchmark & \#train & \#test & words & pairwise sim & hardest sim \\
\midrule
DOCCI & 9450 & 5100 & 123 & 0.84 & 0.93 \\
DCI & 5445 & 1999 & 133 & 0.85 & 0.92 \\
Long-DCI & 5445 (=DCI) & 7444 & 134 & 0.85 & 0.93 \\
Urban-1K & 14579 (VG) & 1000 & 107 & 0.88 & 0.94 \\
\bottomrule
\end{tabular}
}
\end{table}

\subsection{One batch, four benchmarks}
\label{app:data-batch}
\begin{figure}[p]
  \centering
  \includegraphics[width=0.9\textwidth]{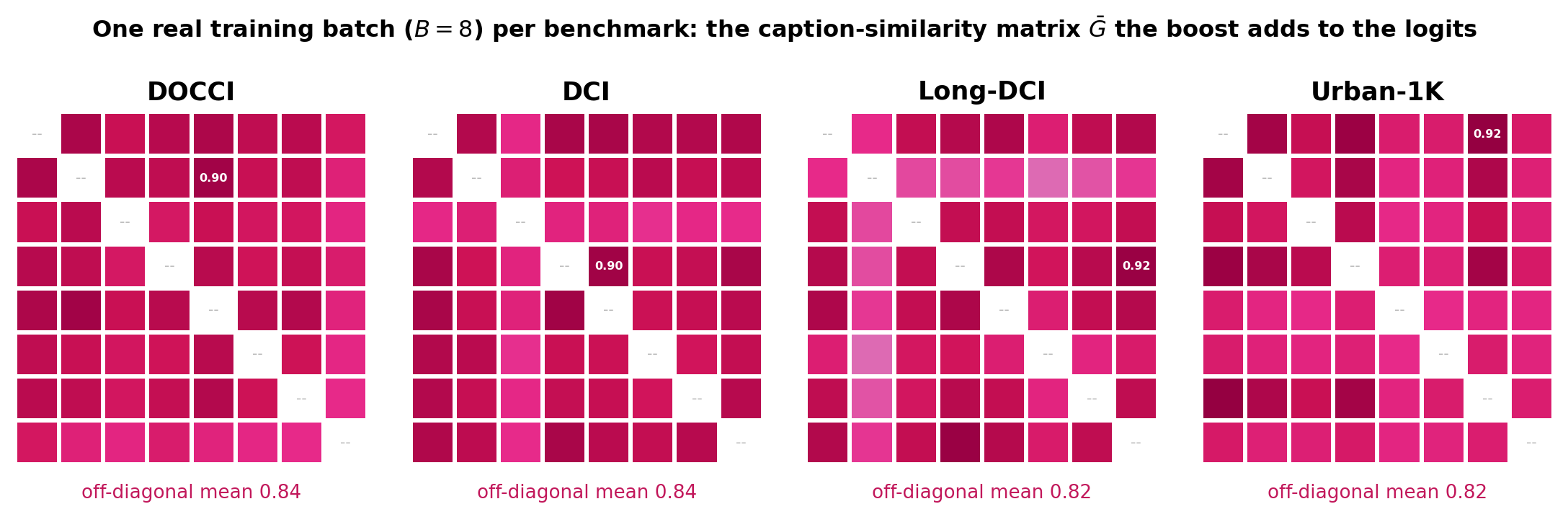}
  \caption{\textbf{One real training batch per benchmark.} Text--text
  similarity matrices $\bar G$ (pre-trained encoder; diagonal masked, max
  off-diagonal entry annotated). Uniformly dark $=$ every negative is
  hard; this matrix, detached and scaled by $\gamma$, is the entire
  mechanism of \method{}.}
  \label{fig:s-batchg}
\end{figure}

 \cref{fig:s-batchg} shows the
matrix at the heart of \method{}: for one \emph{real} training batch
($B{=}8$, same random seed) per benchmark, the caption-similarity matrix
$\bar G$ that our loss adds to the logits. The picture is the method's
motivation made visible: on every benchmark the matrix is uniformly high
(off-diagonal means $0.82$--$0.84$), i.e., \emph{every} negative in
\emph{every} batch is a near-duplicate to some degree, and the boost
assigns every one of them a proportionate margin.

\subsection{What near-duplicates look like}
\label{app:data-neardup}

\begin{figure}[p]
  \centering
  \includegraphics[width=0.92\textwidth]{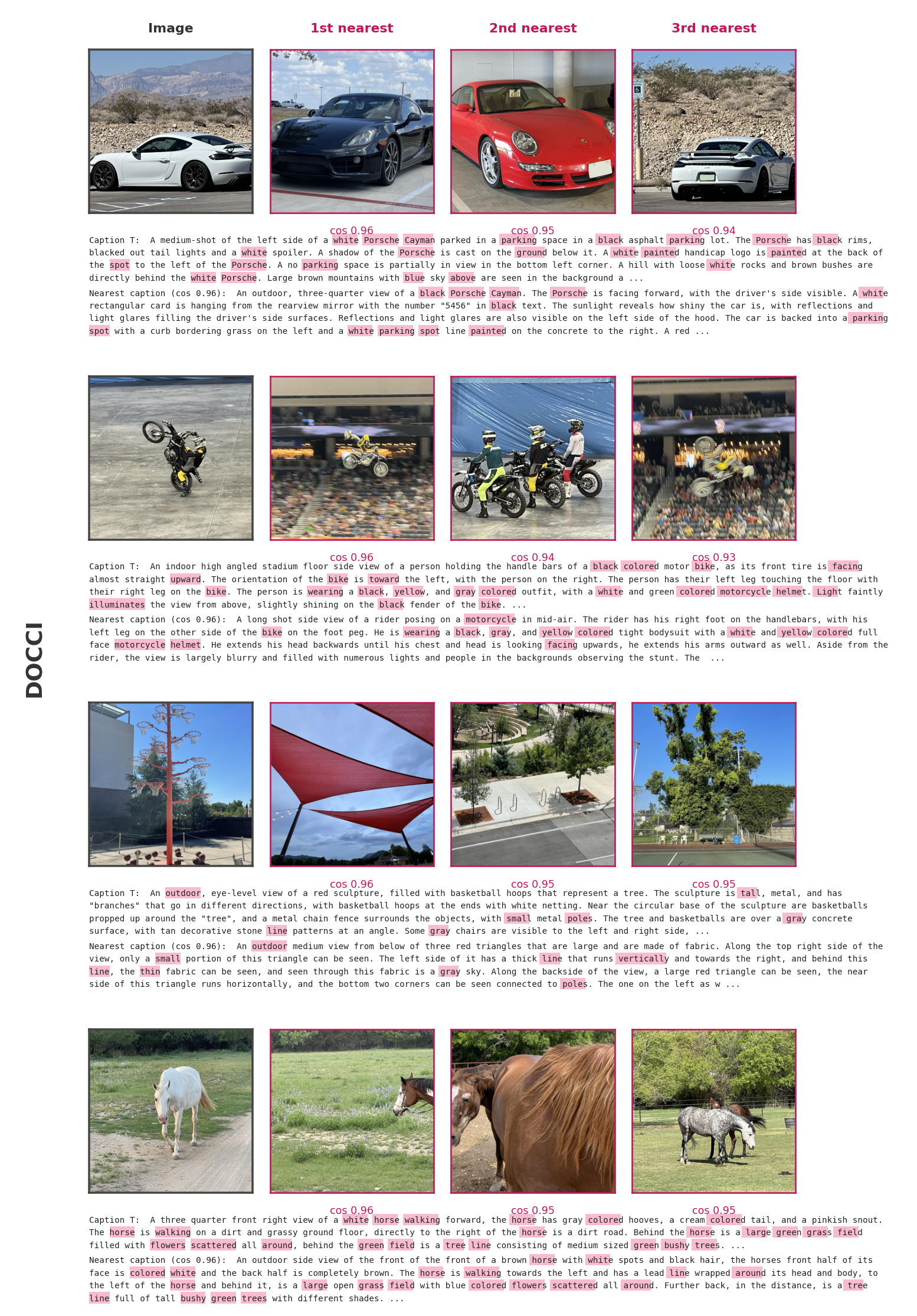}
  \caption{\textbf{Near-duplicate caption pairs on DOCCI.} Each row: a
  test image, its caption, the three nearest \emph{other} captions'
  images (cosine printed underneath), and the caption texts with shared
  content words \colorbox[HTML]{F8BBD0}{highlighted}.}
  \label{fig:s-nd-docci}
\end{figure}

\begin{figure}[p]
  \centering
  \includegraphics[width=0.92\textwidth]{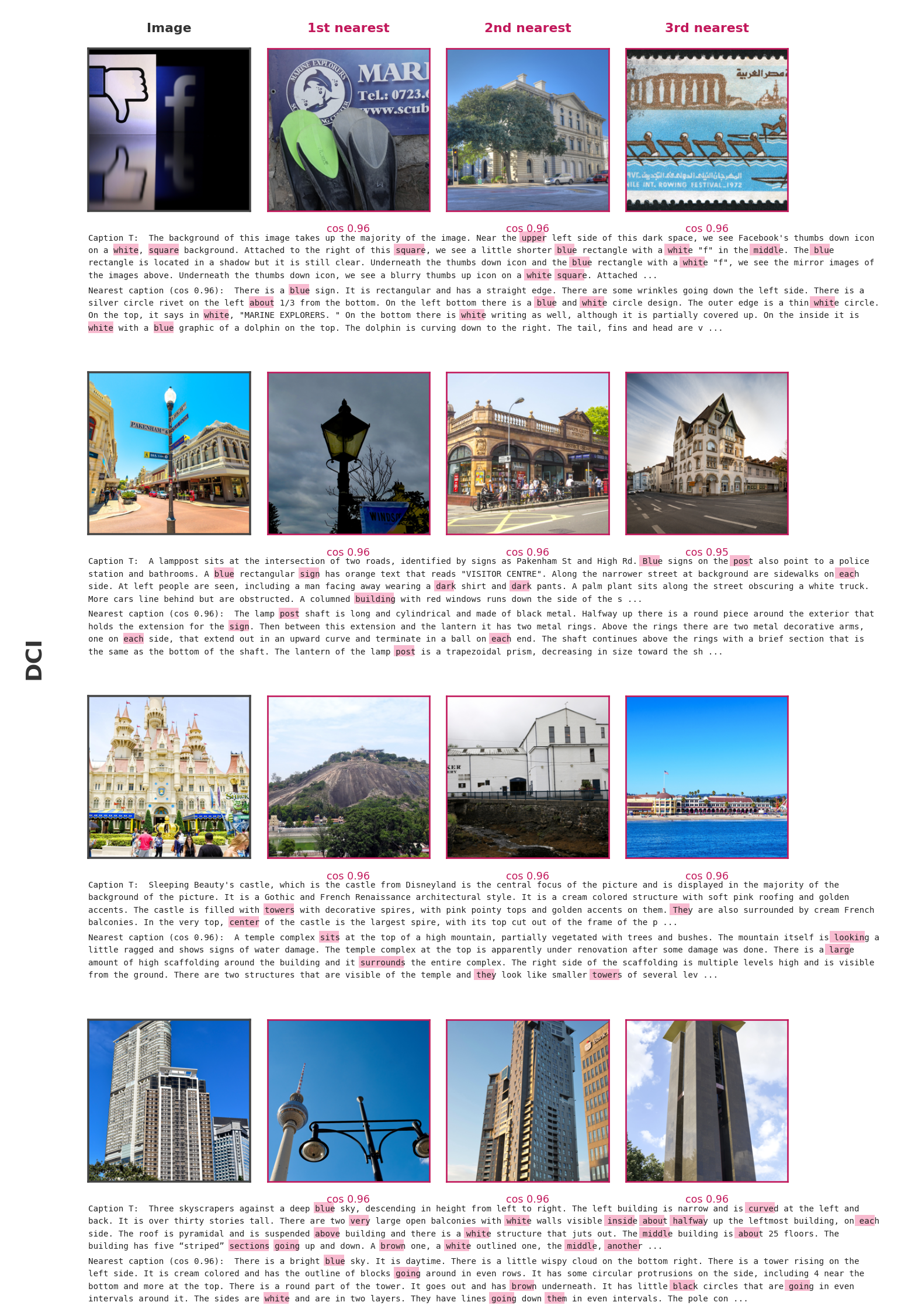}
  \caption{\textbf{Near-duplicate caption pairs on DCI.}}
  \label{fig:s-nd-dci}
\end{figure}

\begin{figure}[p]
  \centering
  \includegraphics[width=0.92\textwidth]{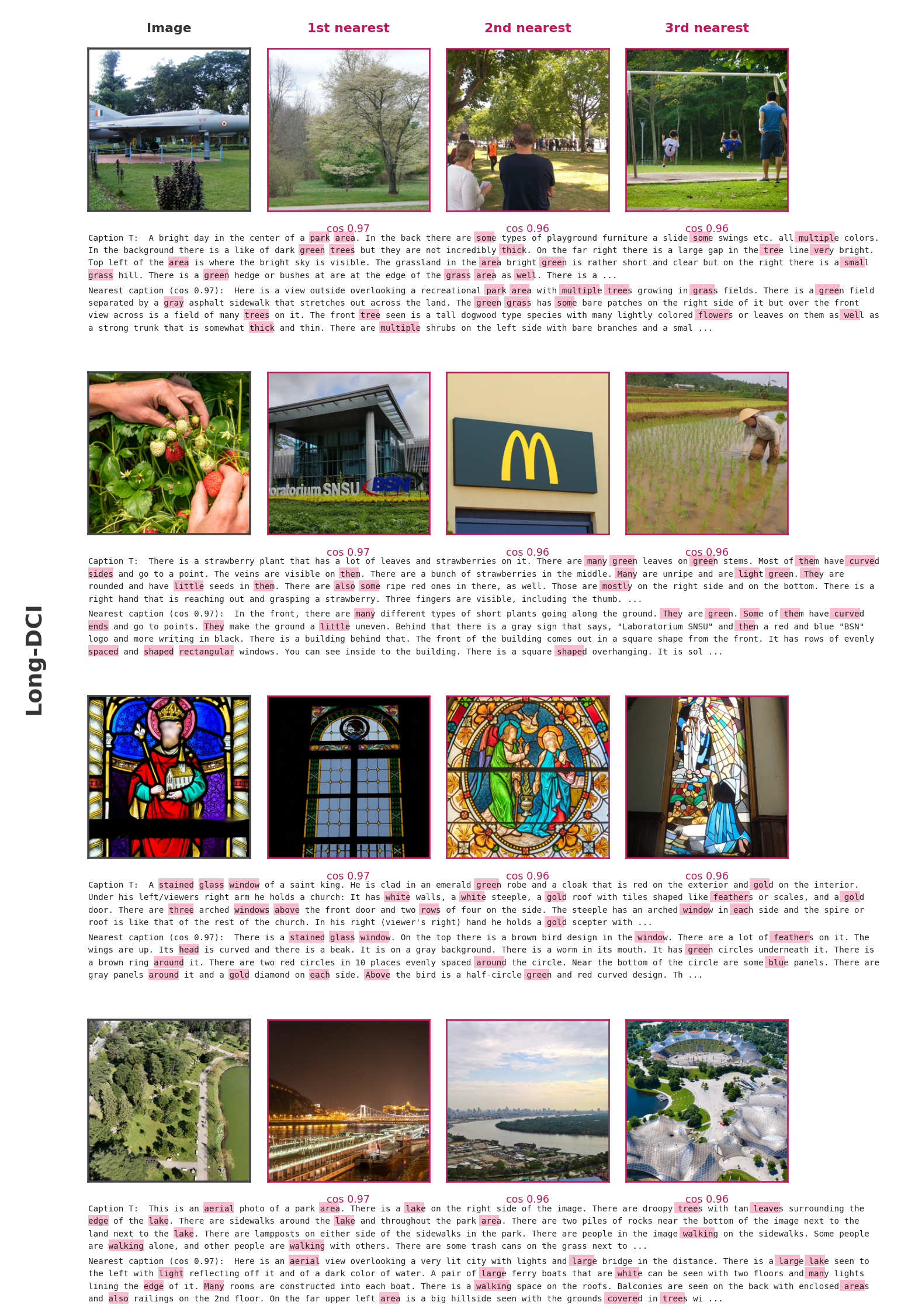}
  \caption{\textbf{Near-duplicate caption pairs on Long-DCI} (full-length
  captions).}
  \label{fig:s-nd-ldci}
\end{figure}

\begin{figure}[p]
  \centering
  \includegraphics[width=0.92\textwidth]{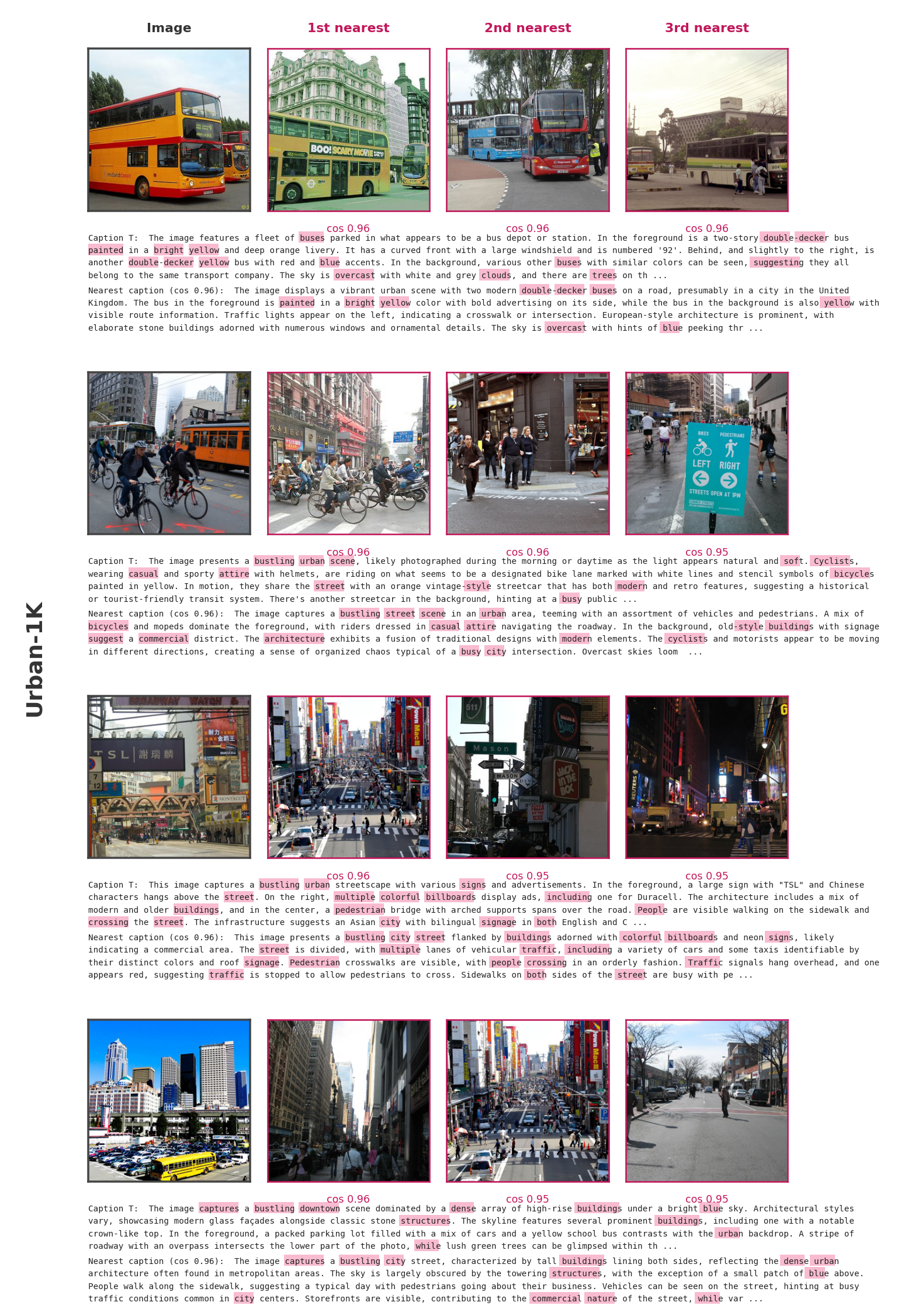}
  \caption{\textbf{Near-duplicate caption pairs on Urban-1K.}}
  \label{fig:s-nd-urban}
\end{figure}

\cref{fig:s-nd-docci,fig:s-nd-dci,fig:s-nd-ldci,fig:s-nd-urban} show
representative near-duplicate caption pairs from each benchmark, retrieved
with the \emph{pre-trained} Long-CLIP-L text encoder (queries drawn near
the 90th percentile of the hardest-companion distribution, i.e., striking
but not extreme). Two different sports cars, two different dirt bikes, two
different white horses: the captions describe \emph{different images} yet
reach cosine similarities of $0.93$--$0.97$, because long descriptions of
natural scenes are compositional recombinations of the same elements
(shared content words highlighted). These are exactly the pairs that
standard InfoNCE treats as ordinary negatives, and the pairs our margin
re-weights hardest.

\subsection{Statistics of the caption geometry}
\label{app:data-stats}
\begin{figure}[p]
  \centering
  \includegraphics[width=0.92\textwidth]{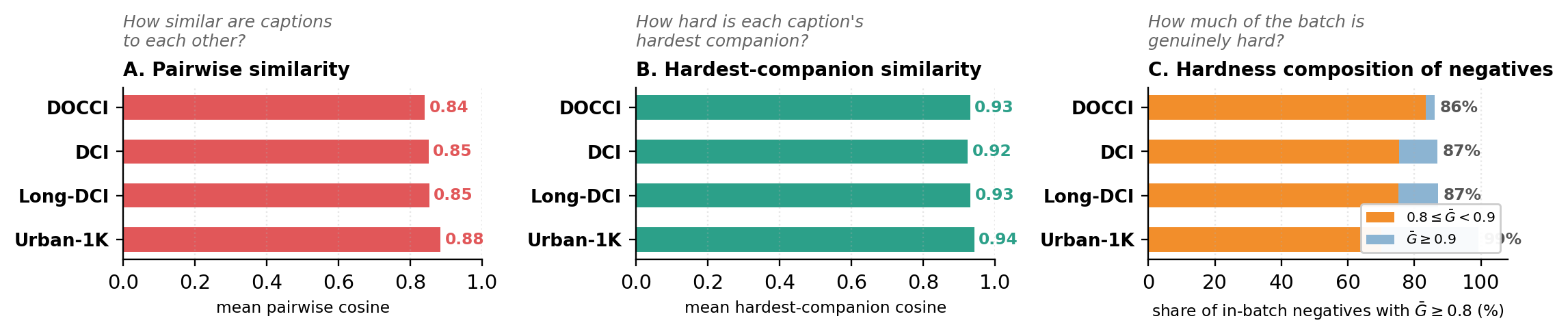}
  \caption{\textbf{Statistical analysis of the caption geometry across the
  four benchmarks} (pre-trained Long-CLIP-L text encoder, full test
  splits). (A)~Mean pairwise caption similarity. (B)~Mean similarity of
  each caption's hardest companion. (C)~Composition of in-batch negatives
  by hardness bucket.}
  \label{fig:s-stats}
\end{figure}

 \cref{fig:s-stats}
quantifies this across the four benchmarks with the same encoder: mean
pairwise similarity is $0.84$--$0.88$ (A), each caption's hardest
companion averages $0.92$--$0.94$ (B), and $86$--$99\%$ of all in-batch
negatives carry a caption similarity of at least $0.8$ (C): under our
default $\gamma{=}0.5$, virtually every negative in every batch receives a
non-trivial adaptive margin, and the benchmarks differ mainly in how many
\emph{extreme} ($\bar G \geq 0.9$) negatives they contain.



\clearpage
\section{Method Details}
\label{app:method-details}

\subsection{Reference implementation}
\label{app:refimpl}
The complete implementation of $\lhn$ is shown in \cref{fig:s-code},
verbatim from our code base; everything else (data loading, optimizer, schedules, evaluation) is standard
Long-CLIP fine-tuning.

\begin{figure}[h]
\centering
\begin{minipage}{0.72\textwidth}
\colorbox{codegray}{\parbox{\dimexpr\linewidth-2\fboxsep}{%
\textcolor{white}{\small\textbf{~The complete implementation of $\lhn$ (PyTorch)}}}}\\[-1pt]
\fcolorbox{codegray}{codebg}{\parbox{\dimexpr\linewidth-2\fboxsep-2\fboxrule}{%
\BUseVerbatim[fontsize=\scriptsize]{hncode}}}
\end{minipage}
\caption{\textbf{Reference implementation.} The complete implementation
of $\lhn$; data loading, optimizer, schedule, and evaluation follow
standard Long-CLIP fine-tuning.}
\label{fig:s-code}
\end{figure}

\subsection{Gradient concentration}
\label{app:deriv}
Why does the boost redirect learning toward hard negatives? Write the
boosted logits as $z_{ij} = s(S_{ij} + \gamma\bar G_{ij})$ and let
$p_{ij} = \mathrm{softmax}_j(z_{i,:})$. The gradient of the cross-entropy
row term w.r.t.\ the similarity $S_{ij}$ of negative $j$ is the classic
softmax residual, $\partial\mathcal{L}_i/\partial S_{ij} = s\,p_{ij}$:
each negative is repelled with force proportional to its posterior. The
boost acts entirely through this posterior. For two negatives $j,k$,
\begin{equation}
\frac{p_{ij}}{p_{ik}}
= \exp\!\big(s\,(S_{ij}-S_{ik})\big)\cdot
  \exp\!\big(s\gamma\,(\bar G_{ij}-\bar G_{ik})\big),
\label{eq:s-conc}
\end{equation}
so relative repulsion is re-weighted by
$\exp(s\gamma\,\Delta\bar G)$: with $s\!\approx\!100$ and $\gamma{=}0.5$,
a caption-similarity gap of just $\Delta\bar G = 0.1$ multiplies the
gradient ratio by $e^{5}\!\approx\!148$. Because $\bar G$ is detached,
the re-weighting is not itself differentiated, so no gradient step is taken
in the direction that would shrink the model's own margins; and because
$\bar G_{ii}=0$, the positive's force $s\,(p_{ii}-1)$ is affected only through the normalizer: the positive must now beat every negative by a
margin $\gamma\bar G_{ij}$ before its own loss vanishes.
\cref{eq:s-conc} is the entire mechanism: hardness enters as an additive
margin, gradients concentrate multiplicatively.

\paragraph{Scope of the stop-gradient, and why the drift is load-bearing.}
\cref{eq:s-conc} also makes explicit what $\mathrm{stopgrad}$ does not
guarantee. We recompute $\bar G$ from the \emph{current} text encoder at
every step, so although no gradient flows through it, $\bar G$ inherits
whatever drift the encoder undergoes under the main objective: the margins
are stationary within a step but not across training. To determine whether
that drift is a defect or a feature, we re-train all three models with
$\bar G$ built instead from a frozen copy of the pre-trained text encoder,
holding seed, schedule, $\gamma$, and every other setting fixed
(\cref{tab:frozeng}).

The drift turns out to be load-bearing. The recomputed default wins all
24 columns, including the average R@1 on all four benchmarks, and, more
diagnostically, the frozen variant's best epoch is the \emph{first} one
on both DCI and Urban-1K, after which accuracy trends downward for the rest
of training. The reason is visible in the magnitudes:
with a mean off-diagonal $\bar G$ of $0.84$ (\cref{fig:s-stats}),
$\gamma{=}0.5$ and $s\!\approx\!100$, a frozen boost imposes a sustained
logit handicap of roughly $42$ that never relaxes, so the model is
permanently over-constrained. Recomputing $\bar G$ lets the boost decay as
training separates the text embeddings, which amounts to an \emph{implicit annealing of the margin}: the objective stops pushing on pairs it has
already resolved. On Long-DCI, the recomputed default leads the frozen variant by
$0.97/1.49$ R@1 in T$\rightarrow$I / I$\rightarrow$T, and by
$2.5$--$3.3$ points at R@5/R@10, showing that a margin that never relaxes
hurts both the top of the ranking and the deeper ranks.

This also corroborates the $\gamma$ sweep from an independent direction.
There, milder boosts transferred better and $\gamma\!\leq\!0.25$ was optimal on
Urban-1K; here, a boost that never decays, effectively the largest
cumulative margin of any setting we ran, is worst on that same transfer
benchmark, trailing the default by $2.90/7.20$ R@1. Two different knobs, the
magnitude of the margin and its schedule, agree that sustained margin
sharpens decision boundaries on the training distribution at the cost of
transfer. We therefore keep the recomputed $\bar G$ throughout, and note
that it is also the cheaper of the two: no second set of weights and no
extra forward pass.

\begin{table}[h]
\centering
\caption{\textbf{Recomputed vs.\ frozen $\bar G$} (Recall@K, \tih{}~and~\ith{}). Building the boost from a frozen copy
of the pre-trained text encoder makes the margins exactly stationary but
removes their implicit annealing: accuracy peaks at epoch~1 on DCI and
Urban-1K and then declines, and the default wins all 24 columns. Best
per column in \textbf{bold}. The final row reports the selected best epoch
as default/frozen.}
\label{tab:frozeng}
\setlength{\tabcolsep}{2.4pt}
\renewcommand{\arraystretch}{1.12}
\scriptsize
\resizebox{\textwidth}{!}{%
\begin{tabular}{l|cccccc|cccccc}
\toprule
& \multicolumn{6}{c|}{\cellcolor{bandgray}\textbf{DOCCI}} & \multicolumn{6}{c}{\cellcolor{bandgray}\textbf{DCI}} \\
$\bar G$ source & \cellcolor{tipink}R@1 & \cellcolor{tipink}R@5 & \cellcolor{tipink}R@10 & \cellcolor{itblue}R@1 & \cellcolor{itblue}R@5 & \cellcolor{itblue}R@10 & \cellcolor{tipink}R@1 & \cellcolor{tipink}R@5 & \cellcolor{tipink}R@10 & \cellcolor{itblue}R@1 & \cellcolor{itblue}R@5 & \cellcolor{itblue}R@10 \\
\midrule
Recomputed $\bar G$ (default) & \textbf{88.25} & \textbf{98.45} & \textbf{99.43} & \textbf{86.24} & \textbf{98.12} & \textbf{99.22} & \textbf{80.69} & \textbf{92.40} & \textbf{95.10} & \textbf{78.84} & \textbf{91.90} & \textbf{94.65} \\
Frozen $\bar G$ & 85.86 & 97.61 & 99.18 & 82.43 & 96.82 & 98.73 & 77.79 & 90.80 & 93.35 & 75.84 & 89.49 & 92.95 \\
\midrule
$\Delta$ (default $-$ frozen) & \up{2.39} & \up{0.84} & \up{0.25} & \up{3.81} & \up{1.30} & \up{0.49} & \up{2.90} & \up{1.60} & \up{1.75} & \up{3.00} & \up{2.41} & \up{1.70} \\
best epoch (default / frozen) & \multicolumn{6}{c|}{6 / 5} & \multicolumn{6}{c}{5 / 1} \\
\midrule
\midrule
& \multicolumn{6}{c|}{\cellcolor{bandgray}\textbf{Long-DCI}} & \multicolumn{6}{c}{\cellcolor{bandgray}\textbf{Urban-1K}} \\
$\bar G$ source & \cellcolor{tipink}R@1 & \cellcolor{tipink}R@5 & \cellcolor{tipink}R@10 & \cellcolor{itblue}R@1 & \cellcolor{itblue}R@5 & \cellcolor{itblue}R@10 & \cellcolor{tipink}R@1 & \cellcolor{tipink}R@5 & \cellcolor{tipink}R@10 & \cellcolor{itblue}R@1 & \cellcolor{itblue}R@5 & \cellcolor{itblue}R@10 \\
\midrule
Recomputed $\bar G$ (default) & \textbf{78.90} & \textbf{93.85} & \textbf{96.34} & \textbf{76.24} & \textbf{92.86} & \textbf{95.90} & \textbf{90.20} & \textbf{98.20} & \textbf{99.20} & \textbf{90.70} & \textbf{98.20} & \textbf{99.30} \\
Frozen $\bar G$ & 77.93 & 90.84 & 93.82 & 74.75 & 89.54 & 92.62 & 87.30 & 97.70 & 98.90 & 83.50 & 95.90 & 97.90 \\
\midrule
$\Delta$ (default $-$ frozen) & \up{0.97} & \up{3.01} & \up{2.52} & \up{1.49} & \up{3.32} & \up{3.28} & \up{2.90} & \up{0.50} & \up{0.30} & \up{7.20} & \up{2.30} & \up{1.40} \\
best epoch (default / frozen) & \multicolumn{6}{c|}{10 / 10} & \multicolumn{6}{c}{4 / 1} \\
\bottomrule
\end{tabular}
}
\end{table}

\paragraph{Interaction with temperature and batch size.} \cref{eq:s-conc}
shows that $s$ and $\gamma$ enter the gradient concentration only through
their product $s\gamma$: changing the inverse temperature rescales the
effective margin, so the broad $\gamma$ plateau of \cref{tab:gamma}a covers
a correspondingly broad range of $s$ at fixed $\gamma$. Batch size acts on
a different axis: because essentially every in-batch negative on these
benchmarks is a near-duplicate (\cref{fig:s-stats}C), enlarging $B$ adds
more boosted pairs rather than diluting them.

\section{Extended Gradient Analysis}
\label{app:gradient}

\paragraph{Reading the median in \cref{fig:gradient}b.} The boosted gradient is
bimodal \emph{within} every epoch: about half of the sampled batches are already
resolved ($\lVert g\rVert<10^{-3}$) while the rest sit at $\mathcal{O}(10^{2})$.
The per-epoch median therefore lands between the two modes and moves with the
sampling mix rather than with a trend. The stable statistic is the upper
quartile, which stays at $\mathcal{O}(10^{2})$ for all ten epochs---including
the last, where the cosine schedule has decayed the learning rate to
$2\times10^{-8}$, showing that hard batches still induce substantial objective
gradients late in training. The standard gradient shows no such structure:
its distribution largely collapses, with the per-epoch median reaching exact
zero in five of the ten epochs.

\begin{figure}[p]
  \centering
  \includegraphics[width=0.85\textwidth]{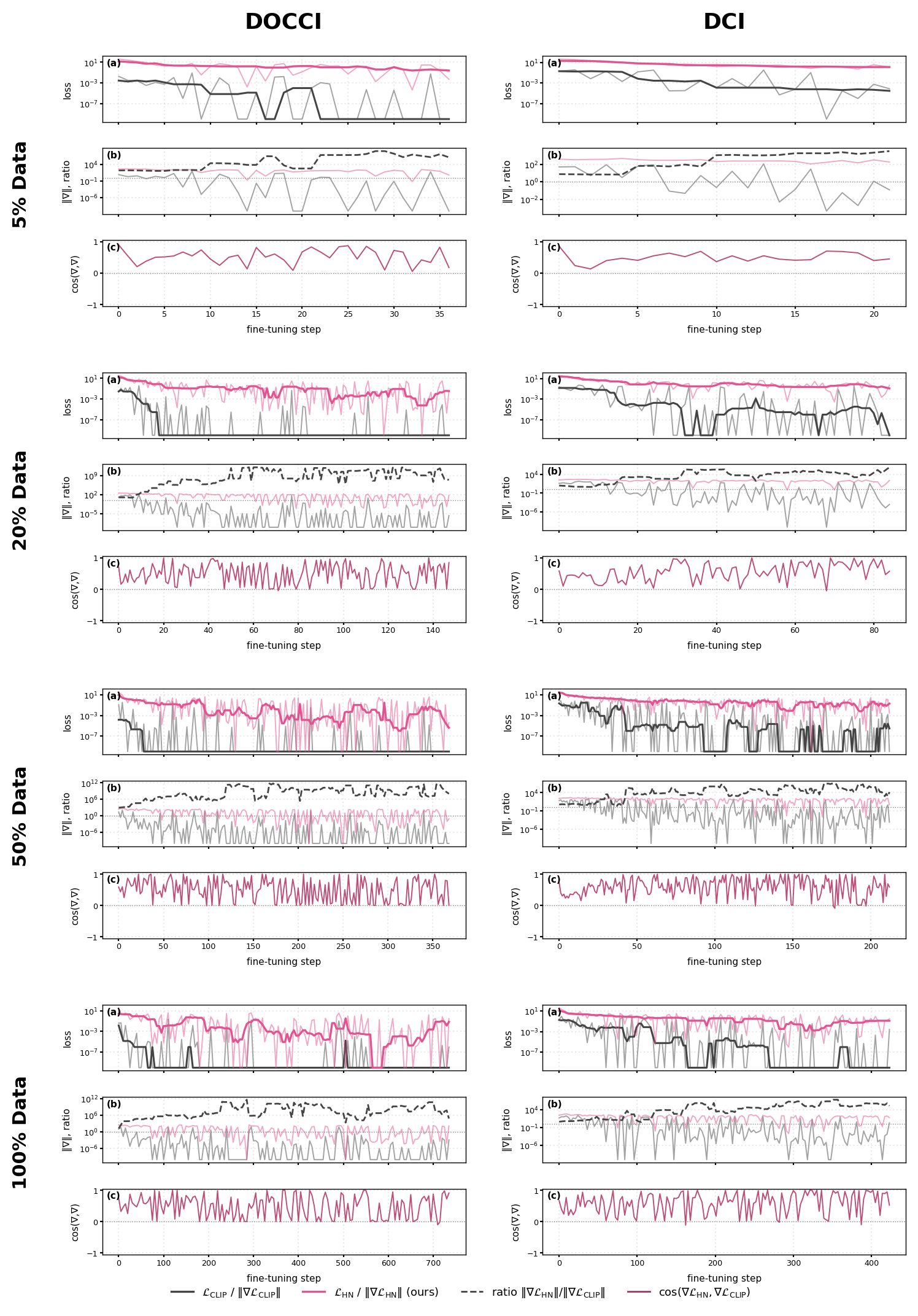}
  \caption{\textbf{Gradient dynamics across data scales.} Rows: fraction
  of the training set; columns: dataset. Per cell: (a)~raw loss traces
  (thin) with running medians (thick), (b)~gradient norms and their ratio
  (dashed), (c)~gradient cosine. Log axes in (a,b); values clamped at
  $10^{-10}$ (numerically zero values).}
  \label{fig:s-grid}
\end{figure}

\cref{fig:s-grid} extends the gradient analysis of \cref{fig:gradient}
across training-data scales: for each fraction of the training set
($5/20/50/100\%$) and each dataset (DOCCI, DCI) we plot the raw traces of
(a)~both losses, (b)~both gradient norms together with their ratio, and
(c)~the cosine between the two gradients. The picture is identical at
every scale: the standard loss collapses to the numerical floor within
the first epoch (gray), the boosted loss keeps producing signal (pink),
their gradient-norm ratio, computed on the $79\%$ of measurements where the
standard gradient is nonzero, runs at a per-run median of $10^{2}$--$10^{6}$
and peaks between $10^{5}$ and $10^{8}$ (dashed), and
the gradient direction remains positively aligned with the standard
objective throughout (median cosine $0.58$, positive on $98\%$ of
measurements, bottom panels), in a
$10^{8}$-dimensional parameter space where random directions are
orthogonal in expectation, this is a strongly compatible update direction.
Saturation is thus not an artifact of one training-set size; it is the
default behavior of InfoNCE on dense captions, and the boost repairs it
wherever it appears.

\section{Implementation Details and Computational Analysis}
\label{app:impl}

\subsection{Training setup and backbone choice}
\label{app:impl-setup}
All methods share the Long-CLIP-L backbone (ViT-L/14, positionally stretched
248-token text encoder), micro-batch 16 with $8\times$ gradient
accumulation (effective batch 128), and the same 10-epoch training budget.
We use a single backbone throughout because the
task requires one: the captions in these benchmarks average $107$--$134$
words on their evaluation splits (\cref{tab:s1}), well beyond the
$77$-token window of standard CLIP, and Long-CLIP-L
is the only publicly released long-context CLIP that every compared method
is also built on. Establishing that the saturation we diagnose is a
property of InfoNCE on dense captions rather than of this particular
checkpoint is therefore left to the caption-length axis, where
\cref{fig:s-grid} shows the same behaviour across four training-set scales
and two datasets, and to the six fine-tuning frameworks of
\cref{tab:s-plug-all}, which differ in architecture and in which
parameters they train. \method{} uses AdamW with learning rate
$2{\times}10^{-6}$ and cosine decay; baselines use their official
hyperparameters (FineLIP: backbone $10^{-6}$, cross-modal module
$2{\times}10^{-4}$; GOAL and StructXLIP: $5{\times}10^{-6}$; LoRA/DoRA
adapters: $5{\times}10^{-5}$, $r{=}16$). \cref{tab:s-eff} reports
measured training efficiency: \method{} trains $2.4\times$ faster than
GOAL and $5.4\times$ faster than StructXLIP on identical hardware while
requiring no auxiliary inputs and no offline preprocessing.

\begin{table}[h]
\centering
\caption{\textbf{Training-efficiency comparison} on DCI fine-tuning (10
epochs, 5.4k images, single Ascend 910B, effective batch 128). Wall-clock
for GOAL/StructXLIP is measured from same-device sequential runs minus
evaluation overhead; FineLIP ran in parallel and is not attributable.
Offline preprocessing time (segmentation, edge extraction, LLM filtering) is
\emph{not} included in the wall-clock.}
\label{tab:s-eff}
\setlength{\tabcolsep}{3pt}
\renewcommand{\arraystretch}{1.15}
\scriptsize
\resizebox{0.8\textwidth}{!}{%
\begin{tabular}{l|cc|cc}
\toprule
Method & Aux.\ training inputs & Offline prep. & Wall-clock & Throughput \\
\midrule
Long-CLIP (plain FT) & none & none & $\approx$16 min & 55.5 img/s \\
FineLIP & none & none & --- & --- \\
GOAL & SAM segments & required & $\approx$41 min & $\approx$22 img/s \\
StructXLIP & edges + LLM lexicon & required & $\approx$92 min & $\approx$10 img/s \\
\textbf{\method{} (ours)} & \textbf{none} & \textbf{none} & \textbf{17 min} & \textbf{53 img/s} \\
\bottomrule
\end{tabular}
}
\end{table}

\subsection{Plug-and-play integration}
\label{app:impl-plug}
 For the plug-and-play study we
replace the \emph{global} contrastive term of each framework with $\lhn$
and change nothing else. Long-CLIP: the symmetric InfoNCE over global
embeddings is swapped directly. FineLIP: the global term is swapped; the
fine-grained cross-modal module and its losses are untouched. GOAL: the
original-pair contrastive term is swapped; segment-level alignment terms
are untouched. StructXLIP: the RGB--caption global term is swapped; the
three structural auxiliary losses are untouched. LoRA/DoRA: adapters
($r{=}16$) are trained with $\lhn$ in place of InfoNCE while the backbone
stays frozen. In every case $\gamma{=}0.5$ without per-framework tuning.

\subsection{Evaluation protocol}
\label{app:impl-eval}
 All numbers in the paper are produced by
one shared evaluation loop: encode the full test split with the released
preprocessing (224 center crop, 248-token captions), L2-normalize, score
by cosine, and report Recall@K for both directions. The same evaluation
procedure is used for all methods. \method{}'s lead is unchanged under
last-epoch evaluation (\cref{sec:exp}), and the Long-DCI R@1 ranking is
preserved under a second random seed (\cref{tab:s-seed}).

\section{Additional Experimental Analyses}
\label{app:analyses}

\cref{tab:s-abl-full,tab:s-gamma-full,tab:s-cross,tab:s-sampeff} extend the
main-text protocol to \emph{deeper ranks} (Recall@\{1, 5, 10, 25, 50\});
the remaining tables in this section report R@1/5/10. In either case, where a configuration overlaps with the main text, its
R@1/5/10 values are bit-identical because it uses the same checkpoint and
evaluation loop.

\subsection{Plug-and-play, DOCCI and Long-DCI}
\label{app:plug}
\cref{tab:s-plug-all} reports the plug-and-play study on DOCCI and
Long-DCI at full width, in the three-row format (framework /
$+\lhn$ / $\Delta$). Average R@1 improves for all twelve
framework--dataset combinations, R@1 improves in 24 of 24 per-direction
cases, and the gains grow with caption length (up to $+23.87$ R@1 for
FineLIP on Long-DCI), which is what a mechanism that sharpens boundaries
between near-duplicate captions predicts.

\begin{table}[p]
\centering
\caption{\textbf{Plug-and-play enhancement of $\lhn$ on CLIP-based
fine-tuning.} Results on DOCCI and Long-DCI for \tih{} and \ith{} retrieval; per framework we report the
official baseline, the same recipe with $\lhn$ replacing its global
InfoNCE term, and the per-column difference $\Delta$ (gain in
\textcolor{deltagreen}{$\uparrow$ green}, drop in
\textcolor{venuegray}{$\downarrow$ gray}; differences within $\pm0.15$ are
shown as \textcolor{venuegray}{$\approx$0}). \textbf{Bold} marks the
$+\lhn$ value where it is the better of the pair.}
\label{tab:s-plug-all}
\setlength{\tabcolsep}{5pt}
\renewcommand{\arraystretch}{1.12}
\scriptsize
\resizebox{\textwidth}{!}{%
\begin{tabular}{l|cccccc|cccccc}
\toprule
& \multicolumn{6}{c|}{\cellcolor{bandgray}\textbf{DOCCI}} & \multicolumn{6}{c}{\cellcolor{bandgray}\textbf{Long-DCI}} \\
Method & \cellcolor{tipink}R@1 & \cellcolor{tipink}R@5 & \cellcolor{tipink}R@10 & \cellcolor{itblue}R@1 & \cellcolor{itblue}R@5 & \cellcolor{itblue}R@10 & \cellcolor{tipink}R@1 & \cellcolor{tipink}R@5 & \cellcolor{tipink}R@10 & \cellcolor{itblue}R@1 & \cellcolor{itblue}R@5 & \cellcolor{itblue}R@10 \\
\midrule
Long-CLIP & 86.45 & 98.00 & 99.31 & 84.10 & 97.84 & 99.04 & 70.24 & 89.62 & 94.00 & 68.44 & 89.35 & 94.67 \\
+our $\lhn$ & \textbf{88.31} & \textbf{98.75} & \textbf{99.45} & \textbf{86.39} & \textbf{98.08} & \textbf{99.20} & \textbf{78.43} & \textbf{92.52} & \textbf{95.27} & \textbf{76.02} & \textbf{91.78} & \textbf{94.76} \\
$\Delta$ & \up{1.86} & \up{0.75} & \textcolor{venuegray}{$\approx$0} & \up{2.29} & \up{0.24} & \up{0.16} & \up{8.19} & \up{2.90} & \up{1.27} & \up{7.58} & \up{2.43} & \textcolor{venuegray}{$\approx$0} \\
\addlinespace[2.5pt]
FineLIP & 77.51 & 96.02 & 98.41 & 69.90 & 93.43 & 97.45 & 59.24 & 77.86 & 83.19 & 49.52 & 75.08 & 82.39 \\
+our $\lhn$ & \textbf{85.47} & \textbf{97.71} & \textbf{99.04} & \textbf{83.88} & \textbf{97.41} & \textbf{98.82} & \textbf{74.92} & \textbf{90.52} & \textbf{93.62} & \textbf{73.39} & \textbf{89.87} & \textbf{93.23} \\
$\Delta$ & \up{7.96} & \up{1.69} & \up{0.63} & \up{13.98} & \up{3.98} & \up{1.37} & \up{15.68} & \up{12.66} & \up{10.43} & \up{23.87} & \up{14.79} & \up{10.84} \\
\addlinespace[2.5pt]
GOAL & 81.53 & 97.02 & 98.80 & 80.86 & 96.24 & 98.63 & 74.11 & 92.73 & 95.75 & 73.29 & 92.18 & 95.78 \\
+our $\lhn$ & \textbf{86.24} & \textbf{98.14} & \textbf{99.35} & \textbf{84.86} & \textbf{97.69} & \textbf{99.20} & \textbf{84.27} & \textbf{94.99} & \textbf{96.57} & \textbf{82.28} & \textbf{94.26} & \textbf{96.18} \\
$\Delta$ & \up{4.71} & \up{1.12} & \up{0.55} & \up{4.00} & \up{1.45} & \up{0.57} & \up{10.16} & \up{2.26} & \up{0.82} & \up{8.99} & \up{2.08} & \up{0.40} \\
\addlinespace[2.5pt]
StructXLIP & 84.73 & 97.69 & 99.00 & 82.61 & 97.08 & 98.71 & 75.31 & 93.15 & 95.92 & 72.56 & 92.65 & 95.75 \\
+our $\lhn$ & \textbf{85.45} & \textbf{97.92} & \textbf{99.12} & \textbf{83.31} & \textbf{97.45} & \textbf{98.92} & \textbf{82.04} & \textbf{94.18} & \textbf{96.23} & \textbf{78.98} & \textbf{93.22} & 95.65 \\
$\Delta$ & \up{0.72} & \up{0.23} & \textcolor{venuegray}{$\approx$0} & \up{0.70} & \up{0.37} & \up{0.21} & \up{6.73} & \up{1.03} & \up{0.31} & \up{6.42} & \up{0.57} & \down{0.10} \\
\addlinespace[2.5pt]
LoRA & 80.49 & 96.16 & 98.55 & 77.96 & 95.80 & 98.04 & 60.08 & 80.68 & 86.86 & 58.97 & 80.02 & 86.53 \\
+our $\lhn$ & \textbf{82.82} & \textbf{97.14} & \textbf{98.86} & \textbf{80.84} & \textbf{96.49} & \textbf{98.63} & \textbf{64.87} & \textbf{83.09} & \textbf{88.35} & \textbf{62.79} & \textbf{81.77} & \textbf{87.31} \\
$\Delta$ & \up{2.33} & \up{0.98} & \up{0.31} & \up{2.88} & \up{0.69} & \up{0.59} & \up{4.79} & \up{2.41} & \up{1.49} & \up{3.82} & \up{1.75} & \up{0.78} \\
\addlinespace[2.5pt]
DoRA & 80.76 & 96.25 & 98.61 & 78.20 & 95.90 & 98.10 & 60.76 & 81.19 & 87.48 & 59.65 & 80.31 & 87.13 \\
+our $\lhn$ & \textbf{83.41} & \textbf{97.33} & \textbf{99.00} & \textbf{81.41} & \textbf{96.65} & \textbf{98.67} & \textbf{65.46} & \textbf{83.57} & \textbf{88.61} & \textbf{63.38} & \textbf{82.15} & \textbf{87.56} \\
$\Delta$ & \up{2.65} & \up{1.08} & \up{0.39} & \up{3.21} & \up{0.75} & \up{0.57} & \up{4.70} & \up{2.38} & \up{1.13} & \up{3.73} & \up{1.84} & \up{0.43} \\
\bottomrule
\end{tabular}
}
\end{table}

\subsection{Full-resolution ablations}
\label{app:ablfull}
\cref{tab:s-abl-full} extends the main-text loss ablation to all four
benchmarks and all ranks, and \cref{tab:s-gamma-full} does the same for
the $\gamma$ sweep. The full objective remains the strongest
configuration overall; on the transfer benchmark the four configurations
compress into a narrow band, with the unboosted variants remaining strongest,
mirroring Appendix~\ref{app:plug}.

\begin{table}[p]
\centering
\caption{\textbf{Loss ablation, full resolution} (all four benchmarks,
all ranks). Best per column in \textbf{bold}.}
\label{tab:s-abl-full}
\setlength{\tabcolsep}{1.6pt}
\renewcommand{\arraystretch}{1.25}
\scriptsize
\resizebox{\textwidth}{!}{%
\begin{tabular}{cc|cccccccccc|cccccccccc}
\toprule
& & \multicolumn{10}{c|}{\cellcolor{bandgray}\textbf{DOCCI}} & \multicolumn{10}{c}{\cellcolor{bandgray}\textbf{DCI}} \\
$\ltok$ & $\lhn$ & \cellcolor{tipink}R@1 & \cellcolor{tipink}R@5 & \cellcolor{tipink}R@10 & \cellcolor{tipink}R@25 & \cellcolor{tipink}R@50 & \cellcolor{itblue}R@1 & \cellcolor{itblue}R@5 & \cellcolor{itblue}R@10 & \cellcolor{itblue}R@25 & \cellcolor{itblue}R@50 & \cellcolor{tipink}R@1 & \cellcolor{tipink}R@5 & \cellcolor{tipink}R@10 & \cellcolor{tipink}R@25 & \cellcolor{tipink}R@50 & \cellcolor{itblue}R@1 & \cellcolor{itblue}R@5 & \cellcolor{itblue}R@10 & \cellcolor{itblue}R@25 & \cellcolor{itblue}R@50 \\
\midrule
\xmark & \xmark & 86.45 & 98.00 & 99.31 & \textbf{99.86} & \textbf{99.98} & 84.10 & 97.84 & 99.04 & 99.78 & \textbf{99.98} & 79.39 & 91.60 & 94.65 & 97.25 & 97.95 & 78.34 & 92.85 & 95.25 & 97.50 & 98.50 \\
\cmark & \xmark & 86.45 & 97.98 & 99.43 & \textbf{99.86} & \textbf{99.98} & 84.29 & 97.90 & 98.94 & 99.78 & 99.94 & 79.24 & 91.55 & 94.25 & 97.00 & 98.00 & 77.54 & \textbf{92.95} & \textbf{95.35} & \textbf{97.55} & \textbf{98.60} \\
\xmark & \cmark & \textbf{88.31} & \textbf{98.75} & \textbf{99.45} & 99.84 & \textbf{99.98} & \textbf{86.39} & 98.08 & 99.20 & 99.82 & 99.96 & \textbf{81.04} & 92.20 & 95.00 & \textbf{97.40} & 98.10 & 78.54 & 91.95 & 94.60 & 97.50 & 98.20 \\
\cmark & \cmark & 88.25 & 98.45 & 99.43 & \textbf{99.86} & \textbf{99.98} & 86.24 & \textbf{98.12} & \textbf{99.22} & \textbf{99.84} & 99.94 & 80.69 & \textbf{92.40} & \textbf{95.10} & 97.35 & \textbf{98.15} & \textbf{78.84} & 91.90 & 94.65 & 97.40 & 98.30 \\
\midrule
\midrule
& & \multicolumn{10}{c|}{\cellcolor{bandgray}\textbf{Long-DCI}} & \multicolumn{10}{c}{\cellcolor{bandgray}\textbf{Urban-1K}} \\
$\ltok$ & $\lhn$ & \cellcolor{tipink}R@1 & \cellcolor{tipink}R@5 & \cellcolor{tipink}R@10 & \cellcolor{tipink}R@25 & \cellcolor{tipink}R@50 & \cellcolor{itblue}R@1 & \cellcolor{itblue}R@5 & \cellcolor{itblue}R@10 & \cellcolor{itblue}R@25 & \cellcolor{itblue}R@50 & \cellcolor{tipink}R@1 & \cellcolor{tipink}R@5 & \cellcolor{tipink}R@10 & \cellcolor{tipink}R@25 & \cellcolor{tipink}R@50 & \cellcolor{itblue}R@1 & \cellcolor{itblue}R@5 & \cellcolor{itblue}R@10 & \cellcolor{itblue}R@25 & \cellcolor{itblue}R@50 \\
\midrule
\xmark & \xmark & 70.24 & 89.62 & 94.00 & \textbf{97.66} & \textbf{98.79} & 68.44 & 89.35 & 94.67 & \textbf{98.11} & \textbf{98.95} & 91.70 & \textbf{99.10} & \textbf{99.50} & \textbf{99.80} & 99.90 & \textbf{93.20} & \textbf{99.00} & \textbf{99.50} & \textbf{99.90} & \textbf{100.00} \\
\cmark & \xmark & 70.96 & 89.52 & 93.95 & 97.29 & 98.63 & 68.08 & 89.24 & 94.25 & 97.77 & 98.94 & \textbf{92.00} & 98.90 & 99.30 & 99.70 & \textbf{100.00} & 92.90 & 98.90 & \textbf{99.50} & 99.80 & 99.90 \\
\xmark & \cmark & 78.43 & 92.52 & 95.27 & 97.21 & 98.47 & 76.02 & 91.78 & 94.76 & 97.31 & 98.29 & 90.60 & 98.30 & 99.40 & 99.60 & 99.80 & 90.20 & 98.30 & 99.40 & 99.80 & 99.80 \\
\cmark & \cmark & \textbf{78.90} & \textbf{93.85} & \textbf{96.34} & 97.26 & 98.56 & \textbf{76.24} & \textbf{92.86} & \textbf{95.90} & 97.42 & 98.47 & 90.20 & 98.20 & 99.20 & 99.60 & 99.80 & 90.70 & 98.20 & 99.30 & 99.70 & 99.90 \\
\end{tabular}
}
\end{table}

\begin{table}[p]
\centering
\caption{\textbf{$\gamma$ sweep, full resolution} (all four benchmarks,
all ranks). Best per column in \textbf{bold}.}
\label{tab:s-gamma-full}
\setlength{\tabcolsep}{1.6pt}
\renewcommand{\arraystretch}{1.25}
\scriptsize
\resizebox{\textwidth}{!}{%
\begin{tabular}{l|cccccccccc|cccccccccc}
\toprule
& \multicolumn{10}{c|}{\cellcolor{bandgray}\textbf{DOCCI}} & \multicolumn{10}{c}{\cellcolor{bandgray}\textbf{DCI}} \\
$\gamma$ & \cellcolor{tipink}R@1 & \cellcolor{tipink}R@5 & \cellcolor{tipink}R@10 & \cellcolor{tipink}R@25 & \cellcolor{tipink}R@50 & \cellcolor{itblue}R@1 & \cellcolor{itblue}R@5 & \cellcolor{itblue}R@10 & \cellcolor{itblue}R@25 & \cellcolor{itblue}R@50 & \cellcolor{tipink}R@1 & \cellcolor{tipink}R@5 & \cellcolor{tipink}R@10 & \cellcolor{tipink}R@25 & \cellcolor{tipink}R@50 & \cellcolor{itblue}R@1 & \cellcolor{itblue}R@5 & \cellcolor{itblue}R@10 & \cellcolor{itblue}R@25 & \cellcolor{itblue}R@50 \\
\midrule
$\gamma{=}0$ & 86.45 & 97.98 & 99.43 & 99.86 & \textbf{99.98} & 84.29 & 97.90 & 98.94 & 99.78 & 99.94 & 79.24 & 91.55 & 94.25 & 97.00 & 98.00 & 77.54 & \textbf{92.95} & 95.35 & 97.55 & \textbf{98.60} \\
$\gamma{=}0.25$ & 87.82 & 98.63 & 99.45 & \textbf{99.88} & \textbf{99.98} & \textbf{86.63} & 98.04 & 99.16 & \textbf{99.84} & 99.96 & \textbf{80.69} & \textbf{92.40} & \textbf{95.35} & \textbf{97.55} & \textbf{98.30} & \textbf{79.19} & 92.85 & \textbf{95.40} & \textbf{97.70} & 98.50 \\
$\gamma{=}0.5$ (default) & \textbf{88.25} & 98.45 & 99.43 & 99.86 & \textbf{99.98} & 86.24 & \textbf{98.12} & 99.22 & \textbf{99.84} & 99.94 & \textbf{80.69} & \textbf{92.40} & 95.10 & 97.35 & 98.15 & 78.84 & 91.90 & 94.65 & 97.40 & 98.30 \\
$\gamma{=}0.75$ & 88.20 & \textbf{98.67} & 99.49 & 99.82 & 99.96 & 85.73 & 98.00 & \textbf{99.25} & 99.78 & \textbf{99.98} & 80.64 & 92.20 & 94.70 & 97.20 & 98.25 & 78.39 & 91.05 & 93.90 & 96.90 & 98.25 \\
$\gamma{=}1.0$ & 87.98 & 98.47 & \textbf{99.51} & 99.82 & 99.96 & 85.02 & 97.69 & 99.02 & 99.76 & 99.96 & 80.04 & 92.00 & 94.40 & 97.10 & \textbf{98.30} & 78.29 & 90.80 & 93.70 & 96.70 & 98.00 \\
\midrule
\midrule
& \multicolumn{10}{c|}{\cellcolor{bandgray}\textbf{Long-DCI}} & \multicolumn{10}{c}{\cellcolor{bandgray}\textbf{Urban-1K}} \\
$\gamma$ & \cellcolor{tipink}R@1 & \cellcolor{tipink}R@5 & \cellcolor{tipink}R@10 & \cellcolor{tipink}R@25 & \cellcolor{tipink}R@50 & \cellcolor{itblue}R@1 & \cellcolor{itblue}R@5 & \cellcolor{itblue}R@10 & \cellcolor{itblue}R@25 & \cellcolor{itblue}R@50 & \cellcolor{tipink}R@1 & \cellcolor{tipink}R@5 & \cellcolor{tipink}R@10 & \cellcolor{tipink}R@25 & \cellcolor{tipink}R@50 & \cellcolor{itblue}R@1 & \cellcolor{itblue}R@5 & \cellcolor{itblue}R@10 & \cellcolor{itblue}R@25 & \cellcolor{itblue}R@50 \\
\midrule
$\gamma{=}0$ & 70.96 & 89.52 & 93.95 & 97.29 & 98.63 & 68.08 & 89.24 & 94.25 & 97.77 & \textbf{98.94} & \textbf{92.00} & \textbf{98.90} & 99.30 & \textbf{99.70} & \textbf{100.00} & 92.90 & \textbf{98.90} & \textbf{99.50} & \textbf{99.80} & \textbf{99.90} \\
$\gamma{=}0.25$ & 76.37 & 92.32 & 95.31 & \textbf{97.65} & \textbf{98.70} & 74.69 & \textbf{92.02} & \textbf{95.55} & \textbf{97.89} & 98.72 & 91.70 & 98.50 & \textbf{99.50} & \textbf{99.70} & 99.80 & \textbf{93.10} & \textbf{98.90} & \textbf{99.50} & 99.70 & 99.80 \\
$\gamma{=}0.5$ (default) & 78.90 & \textbf{93.85} & \textbf{96.34} & 97.26 & 98.56 & 76.24 & \textbf{92.86} & \textbf{95.90} & 97.42 & 98.47 & 90.20 & 98.20 & 99.20 & 99.60 & 99.80 & 90.70 & 98.20 & 99.30 & 99.70 & \textbf{99.90} \\
$\gamma{=}0.75$ & 79.82 & 92.42 & 94.99 & 97.03 & 98.13 & \textbf{76.59} & 91.20 & 94.37 & 96.79 & 97.96 & 88.80 & 98.50 & 99.10 & 99.60 & 99.60 & 89.70 & 97.90 & 98.90 & \textbf{99.80} & \textbf{99.90} \\
$\gamma{=}1.0$ & \textbf{80.05} & 92.14 & 94.64 & 96.78 & 97.89 & 76.48 & 90.89 & 93.95 & 96.47 & 97.68 & 88.20 & 98.20 & 99.20 & 99.50 & 99.60 & 88.30 & 97.50 & 98.60 & \textbf{99.80} & \textbf{99.90} \\
\end{tabular}
}
\end{table}

\begin{table}[p]
\centering
\caption{\textbf{Cross-domain transfer DOCCI$\rightarrow$Long-DCI.} Best
per column in \textbf{bold}.}
\label{tab:s-cross}
\setlength{\tabcolsep}{2.6pt}
\renewcommand{\arraystretch}{1.15}
\scriptsize
\resizebox{0.85\textwidth}{!}{%
\begin{tabular}{l|ccccc|ccccc}
\toprule
& \multicolumn{10}{c}{\cellcolor{bandgray}\textbf{DOCCI $\rightarrow$ Long-DCI}} \\
Method & \cellcolor{tipink}R@1 & \cellcolor{tipink}R@5 & \cellcolor{tipink}R@10 & \cellcolor{tipink}R@25 & \cellcolor{tipink}R@50 & \cellcolor{itblue}R@1 & \cellcolor{itblue}R@5 & \cellcolor{itblue}R@10 & \cellcolor{itblue}R@25 & \cellcolor{itblue}R@50 \\
\midrule
Long-CLIP (zero-shot) & 54.61 & 72.80 & 78.33 & 85.29 & 89.15 & 47.35 & 73.04 & 80.10 & 86.66 & 90.60 \\
FineLIP & 54.54 & 72.93 & 79.42 & 85.87 & 89.98 & 46.44 & 72.96 & 80.47 & 87.04 & 90.68 \\
GOAL & 55.74 & 74.83 & 80.63 & 86.78 & 90.54 & 56.60 & 75.48 & 81.25 & 87.45 & 91.30 \\
StructXLIP & 57.70 & 76.03 & 81.68 & 88.06 & 91.60 & 57.90 & 76.36 & 82.09 & 87.71 & 91.03 \\
\textbf{\method{}} & \textbf{62.60} & \textbf{79.16} & \textbf{84.19} & \textbf{89.88} & \textbf{93.05} & \textbf{63.77} & \textbf{79.89} & \textbf{84.22} & \textbf{89.05} & \textbf{92.14} \\
\bottomrule
\end{tabular}
}
\end{table}

\begin{table}[p]
\centering
\caption{\textbf{Seed replication on Long-DCI.}
$\max|\Delta|$ is the largest absolute difference across the six metrics.}
\label{tab:s-seed}
\setlength{\tabcolsep}{2.2pt}
\renewcommand{\arraystretch}{1.15}
\scriptsize
\resizebox{0.75\textwidth}{!}{%
\begin{tabular}{ll|ccc|ccc|c}
\toprule
Method & Seed & \cellcolor{tipink}R@1 & \cellcolor{tipink}R@5 & \cellcolor{tipink}R@10 & \cellcolor{itblue}R@1 & \cellcolor{itblue}R@5 & \cellcolor{itblue}R@10 & $\max|\Delta|$ \\
\midrule
\method{} (ours) & seed 42 & 78.90 & 93.85 & 96.34 & 76.24 & 92.86 & 95.90 & \\
 & seed 43 & 78.79 & 92.32 & 95.12 & 76.18 & 91.99 & 95.15 & 1.53 \\
\midrule
GOAL & seed 42 & 74.11 & 92.73 & 95.75 & 73.29 & 92.18 & 95.78 & \\
 & seed 43 & 74.34 & 92.89 & 95.77 & 72.03 & 92.37 & 95.69 & 1.26 \\
\midrule
StructXLIP & seed 42 & 75.31 & 93.15 & 95.92 & 72.56 & 92.65 & 95.75 & \\
 & seed 43 & 75.58 & 93.12 & 96.20 & 72.39 & 93.03 & 95.80 & 0.38 \\
\bottomrule
\end{tabular}
}
\end{table}

\begin{table}[p]
\centering
\caption{\textbf{Sample efficiency, full resolution.} Best per pair in
\textbf{bold}; $\Delta$ is the average-R@1 margin of \method{} over GOAL
at that fraction.}
\label{tab:s-sampeff}
\setlength{\tabcolsep}{2.2pt}
\renewcommand{\arraystretch}{1.12}
\scriptsize
\resizebox{0.9\textwidth}{!}{%
\begin{tabular}{ll|ccccc|ccccc|c}
\toprule
Data & Method & \cellcolor{tipink}R@1 & \cellcolor{tipink}R@5 & \cellcolor{tipink}R@10 & \cellcolor{tipink}R@25 & \cellcolor{tipink}R@50 & \cellcolor{itblue}R@1 & \cellcolor{itblue}R@5 & \cellcolor{itblue}R@10 & \cellcolor{itblue}R@25 & \cellcolor{itblue}R@50 & $\Delta$ \\
\midrule
\multicolumn{13}{c}{\cellcolor{bandgray}\textbf{DOCCI}} \\
\midrule
5\% & GOAL & 77.88 & 95.35 & 97.88 & 99.59 & 99.84 & 75.45 & 94.65 & 97.76 & 99.35 & 99.84 & \\
 & \textbf{\method{}} & \textbf{82.73} & \textbf{96.94} & \textbf{98.69} & \textbf{99.69} & \textbf{99.84} & \textbf{80.08} & \textbf{96.22} & \textbf{98.35} & \textbf{99.57} & \textbf{99.98} & \up{4.74} \\
\addlinespace[1.5pt]
20\% & GOAL & 78.22 & 95.65 & 98.22 & 99.55 & 99.86 & 76.47 & 94.88 & 97.96 & 99.39 & 99.84 & \\
 & \textbf{\method{}} & \textbf{85.43} & \textbf{97.86} & \textbf{99.20} & \textbf{99.80} & \textbf{99.96} & \textbf{83.75} & \textbf{97.25} & \textbf{98.80} & \textbf{99.76} & \textbf{99.98} & \up{7.25} \\
\addlinespace[1.5pt]
50\% & GOAL & 80.20 & 96.47 & 98.33 & 99.63 & 99.88 & 78.18 & 95.37 & 97.96 & 99.43 & 99.92 & \\
 & \textbf{\method{}} & \textbf{87.16} & \textbf{98.39} & \textbf{99.43} & \textbf{99.84} & \textbf{99.94} & \textbf{85.37} & \textbf{97.69} & \textbf{99.18} & \textbf{99.80} & \textbf{99.94} & \up{7.07} \\
\addlinespace[1.5pt]
100\% & GOAL & 81.53 & 97.02 & 98.80 & 99.78 & 99.92 & 80.86 & 96.24 & 98.63 & 99.65 & 99.90 & \\
 & \textbf{\method{}} & \textbf{88.25} & \textbf{98.45} & \textbf{99.43} & \textbf{99.86} & \textbf{99.98} & \textbf{86.24} & \textbf{98.12} & \textbf{99.22} & \textbf{99.84} & \textbf{99.94} & \up{5.85} \\
\midrule
\multicolumn{13}{c}{\cellcolor{bandgray}\textbf{DCI}} \\
\midrule
5\% & GOAL & 70.94 & 87.19 & 90.95 & 94.95 & 96.95 & 69.03 & 86.89 & 90.80 & 94.75 & 96.85 & \\
 & \textbf{\method{}} & \textbf{73.59} & \textbf{87.39} & \textbf{92.30} & \textbf{95.10} & \textbf{96.95} & \textbf{74.59} & \textbf{89.19} & \textbf{93.20} & \textbf{96.00} & \textbf{97.15} & \up{4.11} \\
\addlinespace[1.5pt]
20\% & GOAL & 73.69 & 87.99 & 91.40 & 95.40 & 97.10 & 71.39 & 87.24 & 91.55 & 95.45 & 97.30 & \\
 & \textbf{\method{}} & \textbf{78.09} & \textbf{90.60} & \textbf{93.70} & \textbf{96.30} & \textbf{97.65} & \textbf{76.79} & \textbf{90.40} & \textbf{93.55} & \textbf{96.55} & \textbf{97.70} & \up{4.90} \\
\addlinespace[1.5pt]
50\% & GOAL & 75.79 & 89.19 & 93.00 & 95.55 & 96.80 & 73.79 & 88.54 & 92.65 & 95.90 & 97.30 & \\
 & \textbf{\method{}} & \textbf{79.24} & \textbf{91.60} & \textbf{94.55} & \textbf{97.05} & \textbf{97.95} & \textbf{78.44} & \textbf{91.35} & \textbf{94.25} & \textbf{96.70} & \textbf{98.00} & \up{4.05} \\
\addlinespace[1.5pt]
100\% & GOAL & 77.29 & 90.25 & 93.30 & 96.10 & 97.40 & 74.84 & 89.94 & 93.25 & 96.50 & 98.15 & \\
 & \textbf{\method{}} & \textbf{80.69} & \textbf{92.40} & \textbf{95.10} & \textbf{97.35} & \textbf{98.15} & \textbf{78.84} & \textbf{91.90} & \textbf{94.65} & \textbf{97.40} & \textbf{98.30} & \up{3.70} \\
\bottomrule
\end{tabular}
}
\end{table}

\subsection{Cross-domain evaluation}
\label{app:cross}
\cref{tab:cross} reports the DCI$\leftrightarrow$DOCCI cross-domain study
discussed in \cref{sec:exp-transfer}: trained on DCI and transferred to DOCCI,
\method{} exceeds every baseline's \emph{in-domain} DOCCI result, and the
reverse transfer keeps \method{} ahead in all six columns.
\cref{tab:s-cross} evaluates the DOCCI-trained models on Long-DCI, a
harder transfer, since both the caption style and the length change.
\method{} leads every column.

\begin{table}[p]
\centering
\caption{\textbf{Cross-domain generalization between DCI and DOCCI.}
Train on one dataset and test on another. Values are Recall@K (\%), using
\tih{} and \ith{} retrieval. In-domain best in \textbf{\textit{italic
bold}}, cross-domain best in \textbf{bold}.}
\label{tab:cross}
\setlength{\tabcolsep}{2.6pt}
\scriptsize
\resizebox{0.9\textwidth}{!}{%
\begin{tabular}{l|ccc|ccc}
\toprule
Setting & \cellcolor{tipink}R@1 & \cellcolor{tipink}R@5 & \cellcolor{tipink}R@10 & \cellcolor{itblue}R@1 & \cellcolor{itblue}R@5 & \cellcolor{itblue}R@10 \\
\midrule
\rowcolor{bandgray}\multicolumn{7}{c}{\textbf{Train on DCI $\rightarrow$ Test on DCI vs.\ DOCCI}} \\
\midrule
Long-CLIP (DCI$\rightarrow$DCI) & \textit{67.83} & \textit{83.19} & \textit{87.69} & \textit{64.13} & \textit{84.84} & \textit{89.74} \\
Long-CLIP (DCI$\rightarrow$DOCCI) & 78.78 & 95.24 & 98.02 & 66.75 & 91.92 & 96.31 \\
\midrule
FineLIP (DCI$\rightarrow$DCI) & \textit{72.69} & \textit{87.14} & \textit{90.65} & \textit{65.48} & \textit{86.84} & \textit{91.00} \\
FineLIP (DCI$\rightarrow$DOCCI) & 80.69 & 96.47 & 98.53 & 65.63 & 91.51 & 96.31 \\
\midrule
GOAL (DCI$\rightarrow$DCI) & \textit{77.29} & \textit{90.25} & \textit{93.30} & \textit{74.84} & \textit{89.94} & \textit{93.25} \\
GOAL (DCI$\rightarrow$DOCCI) & 80.14 & 96.25 & 98.39 & 77.24 & 95.22 & 97.90 \\
\midrule
StructXLIP (DCI$\rightarrow$DCI) & \textit{75.84} & \textit{89.94} & \textit{93.65} & \textit{74.49} & \textit{90.05} & \textit{93.40} \\
StructXLIP (DCI$\rightarrow$DOCCI) & 76.94 & 95.06 & 97.76 & 74.24 & 94.12 & 97.51 \\
\midrule
\textbf{\method{}} (DCI$\rightarrow$DCI) & \textbf{\textit{80.69}} & \textbf{\textit{92.40}} & \textbf{\textit{95.10}} & \textbf{\textit{78.84}} & \textbf{\textit{91.90}} & \textbf{\textit{94.65}} \\
\textbf{\method{}} (DCI$\rightarrow$DOCCI) & \textbf{85.00} & \textbf{97.90} & \textbf{99.25} & \textbf{83.14} & \textbf{96.98} & \textbf{98.75} \\
\midrule
\rowcolor{bandgray}\multicolumn{7}{c}{\textbf{Train on DOCCI $\rightarrow$ Test on DOCCI vs.\ DCI}} \\
\midrule
Long-CLIP (DOCCI$\rightarrow$DOCCI) & \textit{78.78} & \textit{95.24} & \textit{98.02} & \textit{66.75} & \textit{91.92} & \textit{96.31} \\
Long-CLIP (DOCCI$\rightarrow$DCI) & 67.83 & 83.19 & 87.69 & 64.13 & 84.84 & 89.74 \\
\midrule
FineLIP (DOCCI$\rightarrow$DOCCI) & \textit{77.51} & \textit{96.02} & \textit{98.41} & \textit{69.90} & \textit{93.43} & \textit{97.45} \\
FineLIP (DOCCI$\rightarrow$DCI) & 68.23 & 83.84 & 88.74 & 63.48 & 85.79 & 89.84 \\
\midrule
GOAL (DOCCI$\rightarrow$DOCCI) & \textit{81.53} & \textit{97.02} & \textit{98.80} & \textit{80.86} & \textit{96.24} & \textit{98.63} \\
GOAL (DOCCI$\rightarrow$DCI) & 69.58 & 85.24 & 89.39 & 69.88 & 85.34 & 89.79 \\
\midrule
StructXLIP (DOCCI$\rightarrow$DOCCI) & \textit{84.73} & \textit{97.69} & \textit{99.00} & \textit{82.61} & \textit{97.08} & \textit{98.71} \\
StructXLIP (DOCCI$\rightarrow$DCI) & 69.93 & 86.74 & 91.05 & 71.39 & 86.99 & 90.60 \\
\midrule
\textbf{\method{}} (DOCCI$\rightarrow$DOCCI) & \textbf{\textit{88.25}} & \textbf{\textit{98.45}} & \textbf{\textit{99.43}} & \textbf{\textit{86.24}} & \textbf{\textit{98.12}} & \textbf{\textit{99.22}} \\
\textbf{\method{}} (DOCCI$\rightarrow$DCI) & \textbf{74.84} & \textbf{88.54} & \textbf{92.40} & \textbf{75.64} & \textbf{88.04} & \textbf{91.35} \\
\bottomrule
\end{tabular}
%
}
\end{table}

\subsection{Robustness to the random seed}
\label{app:seed}
\cref{tab:s-seeds3} repeats the main comparison under \emph{three} independent
seeds (42, 1337, 3407) and reports mean\,$\pm$\,standard deviation, matching the
protocol StructXLIP uses in its appendix, so the two papers' variability figures
are directly comparable. The ordering of \cref{tab:main} is reproduced under
replication: \method{} holds the best mean in all $24$ columns, including every
R@1 column on all four benchmarks, by $+2.74$ to $+4.08$. The standard
deviations are small---HN-CLIP's R@1 standard deviations range from
$0.14$ to $0.83$, remaining well below its margins---so the R@1 result is
not seed-sensitive. The Long-DCI R@5/R@10
columns also favor \method{} in both directions, so the seed-level advantage
is not confined to R@1.

\begin{table}[t]
\centering
\small
\caption{\textbf{Three-seed replication} (seeds 42, 1337, 3407), mean\,$\pm$\,std
over the three runs; the epoch is selected per benchmark by the same rule used
throughout. \textbf{Bold} marks the best mean per column, \underline{underline}
the runner-up. Long-CLIP is evaluated zero-shot without fine-tuning and
therefore carries no seed variance.}
\label{tab:s-seeds3}
\resizebox{\textwidth}{!}{%
\begin{tabular}{l|cccccc}
\toprule
\multicolumn{7}{c}{\cellcolor{bandgray}\textbf{DOCCI}} \\
Method & \cellcolor{tipink}R@1 & \cellcolor{tipink}R@5 & \cellcolor{tipink}R@10 & \cellcolor{itblue}R@1 & \cellcolor{itblue}R@5 & \cellcolor{itblue}R@10 \\
\midrule
Long-CLIP\vtag{ECCV'24} (zero-shot) & 78.78 & 95.24 & 98.02 & 66.75 & 91.92 & 96.31 \\
FineLIP\vtag{CVPR'25} & 77.39\,{\scriptsize\textcolor{venuegray}{$\pm$0.33}} & 95.82\,{\scriptsize\textcolor{venuegray}{$\pm$0.08}} & 98.36\,{\scriptsize\textcolor{venuegray}{$\pm$0.02}} & 69.80\,{\scriptsize\textcolor{venuegray}{$\pm$0.08}} & 93.14\,{\scriptsize\textcolor{venuegray}{$\pm$0.14}} & 97.37\,{\scriptsize\textcolor{venuegray}{$\pm$0.09}} \\
GOAL\vtag{CVPR'25} & 81.89\,{\scriptsize\textcolor{venuegray}{$\pm$0.36}} & 96.97\,{\scriptsize\textcolor{venuegray}{$\pm$0.04}} & 98.78\,{\scriptsize\textcolor{venuegray}{$\pm$0.05}} & 80.59\,{\scriptsize\textcolor{venuegray}{$\pm$0.51}} & 96.19\,{\scriptsize\textcolor{venuegray}{$\pm$0.10}} & 98.59\,{\scriptsize\textcolor{venuegray}{$\pm$0.07}} \\
StructXLIP\vtag{CVPR'26} & \underline{84.73}\,{\scriptsize\textcolor{venuegray}{$\pm$0.09}} & \underline{97.64}\,{\scriptsize\textcolor{venuegray}{$\pm$0.05}} & \underline{99.04}\,{\scriptsize\textcolor{venuegray}{$\pm$0.09}} & \underline{82.52}\,{\scriptsize\textcolor{venuegray}{$\pm$0.17}} & \underline{96.97}\,{\scriptsize\textcolor{venuegray}{$\pm$0.09}} & \underline{98.78}\,{\scriptsize\textcolor{venuegray}{$\pm$0.08}} \\
\textbf{\method{}} & \textbf{88.36}\,{\scriptsize\textcolor{venuegray}{$\pm$0.14}} & \textbf{98.50}\,{\scriptsize\textcolor{venuegray}{$\pm$0.04}} & \textbf{99.48}\,{\scriptsize\textcolor{venuegray}{$\pm$0.05}} & \textbf{86.41}\,{\scriptsize\textcolor{venuegray}{$\pm$0.19}} & \textbf{98.09}\,{\scriptsize\textcolor{venuegray}{$\pm$0.03}} & \textbf{99.24}\,{\scriptsize\textcolor{venuegray}{$\pm$0.02}} \\
\cmidrule(lr){1-7}
$\Delta$ & \up{3.62} & \up{0.86} & \up{0.44} & \up{3.89} & \up{1.12} & \up{0.46} \\
\midrule\midrule
\multicolumn{7}{c}{\cellcolor{bandgray}\textbf{DCI}} \\
Method & \cellcolor{tipink}R@1 & \cellcolor{tipink}R@5 & \cellcolor{tipink}R@10 & \cellcolor{itblue}R@1 & \cellcolor{itblue}R@5 & \cellcolor{itblue}R@10 \\
\midrule
Long-CLIP\vtag{ECCV'24} (zero-shot) & 67.83 & 83.19 & 87.69 & 64.13 & 84.84 & 89.74 \\
FineLIP\vtag{CVPR'25} & 72.77\,{\scriptsize\textcolor{venuegray}{$\pm$0.21}} & 87.22\,{\scriptsize\textcolor{venuegray}{$\pm$0.03}} & 90.92\,{\scriptsize\textcolor{venuegray}{$\pm$0.08}} & 65.55\,{\scriptsize\textcolor{venuegray}{$\pm$0.35}} & 86.92\,{\scriptsize\textcolor{venuegray}{$\pm$0.06}} & 90.90\,{\scriptsize\textcolor{venuegray}{$\pm$0.10}} \\
GOAL\vtag{CVPR'25} & \underline{76.96}\,{\scriptsize\textcolor{venuegray}{$\pm$0.49}} & 90.03\,{\scriptsize\textcolor{venuegray}{$\pm$0.20}} & 93.23\,{\scriptsize\textcolor{venuegray}{$\pm$0.26}} & \underline{75.07}\,{\scriptsize\textcolor{venuegray}{$\pm$0.32}} & 89.72\,{\scriptsize\textcolor{venuegray}{$\pm$0.20}} & 93.32\,{\scriptsize\textcolor{venuegray}{$\pm$0.06}} \\
StructXLIP\vtag{CVPR'26} & 76.56\,{\scriptsize\textcolor{venuegray}{$\pm$0.66}} & \underline{90.05}\,{\scriptsize\textcolor{venuegray}{$\pm$0.14}} & \underline{93.52}\,{\scriptsize\textcolor{venuegray}{$\pm$0.12}} & 74.42\,{\scriptsize\textcolor{venuegray}{$\pm$0.12}} & \underline{90.37}\,{\scriptsize\textcolor{venuegray}{$\pm$0.30}} & \underline{93.68}\,{\scriptsize\textcolor{venuegray}{$\pm$0.26}} \\
\textbf{\method{}} & \textbf{80.79}\,{\scriptsize\textcolor{venuegray}{$\pm$0.17}} & \textbf{92.22}\,{\scriptsize\textcolor{venuegray}{$\pm$0.16}} & \textbf{94.80}\,{\scriptsize\textcolor{venuegray}{$\pm$0.26}} & \textbf{79.16}\,{\scriptsize\textcolor{venuegray}{$\pm$0.35}} & \textbf{91.93}\,{\scriptsize\textcolor{venuegray}{$\pm$0.15}} & \textbf{94.55}\,{\scriptsize\textcolor{venuegray}{$\pm$0.10}} \\
\cmidrule(lr){1-7}
$\Delta$ & \up{3.83} & \up{2.17} & \up{1.28} & \up{4.08} & \up{1.57} & \up{0.87} \\
\midrule\midrule
\multicolumn{7}{c}{\cellcolor{bandgray}\textbf{Long-DCI}} \\
Method & \cellcolor{tipink}R@1 & \cellcolor{tipink}R@5 & \cellcolor{tipink}R@10 & \cellcolor{itblue}R@1 & \cellcolor{itblue}R@5 & \cellcolor{itblue}R@10 \\
\midrule
Long-CLIP\vtag{ECCV'24} (zero-shot) & 54.61 & 72.80 & 78.33 & 47.35 & 73.04 & 80.10 \\
FineLIP\vtag{CVPR'25} & 59.38\,{\scriptsize\textcolor{venuegray}{$\pm$0.14}} & 77.85\,{\scriptsize\textcolor{venuegray}{$\pm$0.08}} & 83.35\,{\scriptsize\textcolor{venuegray}{$\pm$0.12}} & 49.20\,{\scriptsize\textcolor{venuegray}{$\pm$0.25}} & 75.10\,{\scriptsize\textcolor{venuegray}{$\pm$0.10}} & 82.33\,{\scriptsize\textcolor{venuegray}{$\pm$0.08}} \\
GOAL\vtag{CVPR'25} & 74.58\,{\scriptsize\textcolor{venuegray}{$\pm$0.50}} & 92.90\,{\scriptsize\textcolor{venuegray}{$\pm$0.20}} & 95.83\,{\scriptsize\textcolor{venuegray}{$\pm$0.09}} & \underline{73.40}\,{\scriptsize\textcolor{venuegray}{$\pm$0.46}} & 92.40\,{\scriptsize\textcolor{venuegray}{$\pm$0.22}} & 95.74\,{\scriptsize\textcolor{venuegray}{$\pm$0.10}} \\
StructXLIP\vtag{CVPR'26} & \underline{75.51}\,{\scriptsize\textcolor{venuegray}{$\pm$0.21}} & \underline{93.23}\,{\scriptsize\textcolor{venuegray}{$\pm$0.07}} & \underline{96.04}\,{\scriptsize\textcolor{venuegray}{$\pm$0.11}} & 73.05\,{\scriptsize\textcolor{venuegray}{$\pm$0.50}} & \underline{92.78}\,{\scriptsize\textcolor{venuegray}{$\pm$0.11}} & \underline{95.89}\,{\scriptsize\textcolor{venuegray}{$\pm$0.18}} \\
\textbf{\method{}} & \textbf{78.84}\,{\scriptsize\textcolor{venuegray}{$\pm$0.15}} & \textbf{93.58}\,{\scriptsize\textcolor{venuegray}{$\pm$0.23}} & \textbf{96.21}\,{\scriptsize\textcolor{venuegray}{$\pm$0.12}} & \textbf{76.14}\,{\scriptsize\textcolor{venuegray}{$\pm$0.30}} & \textbf{92.83}\,{\scriptsize\textcolor{venuegray}{$\pm$0.06}} & \textbf{96.01}\,{\scriptsize\textcolor{venuegray}{$\pm$0.11}} \\
\cmidrule(lr){1-7}
$\Delta$ & \up{3.33} & \up{0.35} & \up{0.17} & \up{2.74} & \up{0.05} & \up{0.12} \\
\midrule\midrule
\multicolumn{7}{c}{\cellcolor{bandgray}\textbf{Urban-1K}} \\
Method & \cellcolor{tipink}R@1 & \cellcolor{tipink}R@5 & \cellcolor{tipink}R@10 & \cellcolor{itblue}R@1 & \cellcolor{itblue}R@5 & \cellcolor{itblue}R@10 \\
\midrule
Long-CLIP\vtag{ECCV'24} (zero-shot) & 86.10 & 96.50 & 98.10 & 82.40 & 96.70 & 98.30 \\
FineLIP\vtag{CVPR'25} & 81.57\,{\scriptsize\textcolor{venuegray}{$\pm$0.35}} & 94.93\,{\scriptsize\textcolor{venuegray}{$\pm$0.15}} & 97.23\,{\scriptsize\textcolor{venuegray}{$\pm$0.21}} & 77.13\,{\scriptsize\textcolor{venuegray}{$\pm$0.99}} & 94.90\,{\scriptsize\textcolor{venuegray}{$\pm$0.10}} & 97.50\,{\scriptsize\textcolor{venuegray}{$\pm$0.10}} \\
GOAL\vtag{CVPR'25} & \underline{86.10}\,{\scriptsize\textcolor{venuegray}{$\pm$1.49}} & \underline{97.00}\,{\scriptsize\textcolor{venuegray}{$\pm$0.53}} & \underline{98.83}\,{\scriptsize\textcolor{venuegray}{$\pm$0.40}} & \underline{88.10}\,{\scriptsize\textcolor{venuegray}{$\pm$0.17}} & \underline{97.10}\,{\scriptsize\textcolor{venuegray}{$\pm$0.36}} & \underline{98.73}\,{\scriptsize\textcolor{venuegray}{$\pm$0.32}} \\
StructXLIP\vtag{CVPR'26} & 85.47\,{\scriptsize\textcolor{venuegray}{$\pm$1.30}} & 96.67\,{\scriptsize\textcolor{venuegray}{$\pm$0.86}} & 98.60\,{\scriptsize\textcolor{venuegray}{$\pm$0.30}} & 87.23\,{\scriptsize\textcolor{venuegray}{$\pm$0.90}} & 96.57\,{\scriptsize\textcolor{venuegray}{$\pm$0.42}} & 98.20\,{\scriptsize\textcolor{venuegray}{$\pm$0.17}} \\
\textbf{\method{}} & \textbf{89.87}\,{\scriptsize\textcolor{venuegray}{$\pm$0.31}} & \textbf{98.37}\,{\scriptsize\textcolor{venuegray}{$\pm$0.15}} & \textbf{99.33}\,{\scriptsize\textcolor{venuegray}{$\pm$0.12}} & \textbf{91.37}\,{\scriptsize\textcolor{venuegray}{$\pm$0.83}} & \textbf{98.57}\,{\scriptsize\textcolor{venuegray}{$\pm$0.35}} & \textbf{99.27}\,{\scriptsize\textcolor{venuegray}{$\pm$0.06}} \\
\cmidrule(lr){1-7}
$\Delta$ & \up{3.77} & \up{1.37} & \up{0.50} & \up{3.27} & \up{1.47} & \up{0.53} \\
\bottomrule
\end{tabular}
}
\end{table}

\cref{tab:s-seed} re-trains \method{}, GOAL, and StructXLIP with a second
random seed under the identical recipe and evaluates on Long-DCI. For
\method{}, the two R@1 values differ by only $0.09$ on average, and the
R@1 ranking is preserved under the second seed (largest R@1 deviation
$1.26$, on GOAL). Across all six metrics, the maximum seed-to-seed
difference is $1.53$ for \method{}, reflecting modest variation at deeper
ranks.

\subsection{Sample efficiency, full numbers}
\label{app:sampeff}
\cref{tab:s-sampeff} lists all measurements behind \cref{fig:sampeff-main}:
both benchmarks, both methods, all fractions and ranks, on identical
training subsets. \method{} leads or ties every cell; the average-R@1
margin (last column) \emph{widens} as data shrinks to $20\%$ on both
benchmarks, and the $20\%$ rows exceed GOAL's $100\%$ rows on both. The
dotted reference line of \cref{fig:sampeff-main} is instead the
\emph{strongest} baseline at $100\%$ (StructXLIP on DOCCI, GOAL on DCI;
\cref{tab:main}), which the $20\%$ rows also clear.

\subsection{Per-direction convergence}
\label{app:conv}
\begin{figure}[p]
  \centering
  \includegraphics[width=\textwidth]{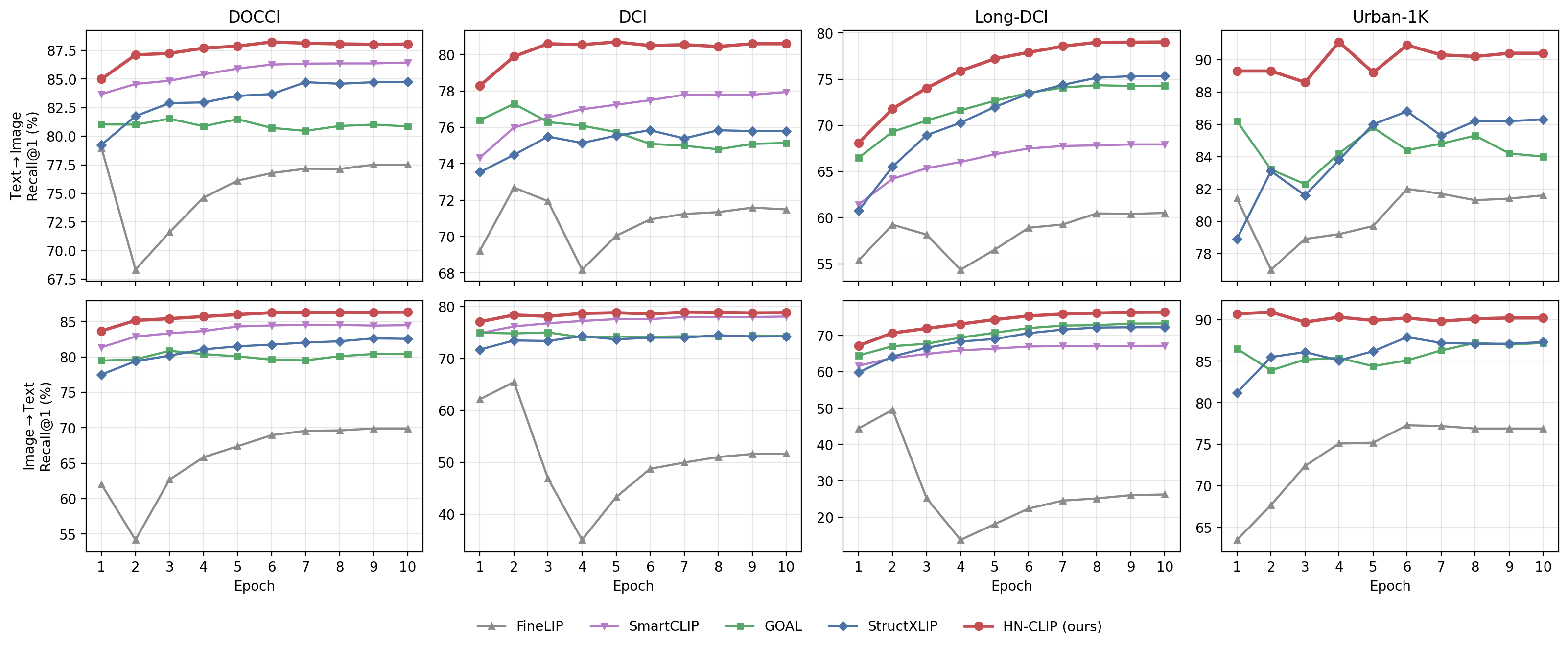}
  \caption{\textbf{Per-direction convergence.} Recall@1 per epoch; top:
  Text$\rightarrow$Image, bottom: Image$\rightarrow$Text.}
  \label{fig:s-conv}
\end{figure}

\cref{fig:s-conv} splits \cref{fig:convergence} by retrieval direction:
\method{} leads in \emph{both} directions on all four benchmarks from the
first epoch, so the averaged curves hide no asymmetry. The split also
exposes training instabilities of FineLIP (e.g., an I$\rightarrow$T
collapse on DCI at epoch~4) that the averaged view smooths over.

\subsection{Results at deeper ranks}
\label{app:deepranks}
At R@5/R@10 the fine-tuned methods approach ceiling on DOCCI and
Urban-1K and margins compress there (\cref{tab:main}); on DCI and
Long-DCI the spread stays wide at every rank. On Long-DCI,
\method{} is best in all four R@5/R@10 columns, while continuing
to lead all eight R@1 columns. The second seed preserves the R@1 ordering
but shows modest variation at deeper ranks. The two mechanisms are
complementary rather than competing:
\cref{tab:plug} shows that adding $\lhn$ \emph{inside} StructXLIP
improves it further.

\section{Additional Qualitative Results}
\label{app:qual}
\begin{figure}[p]
  \centering
  \includegraphics[width=0.85\textwidth]{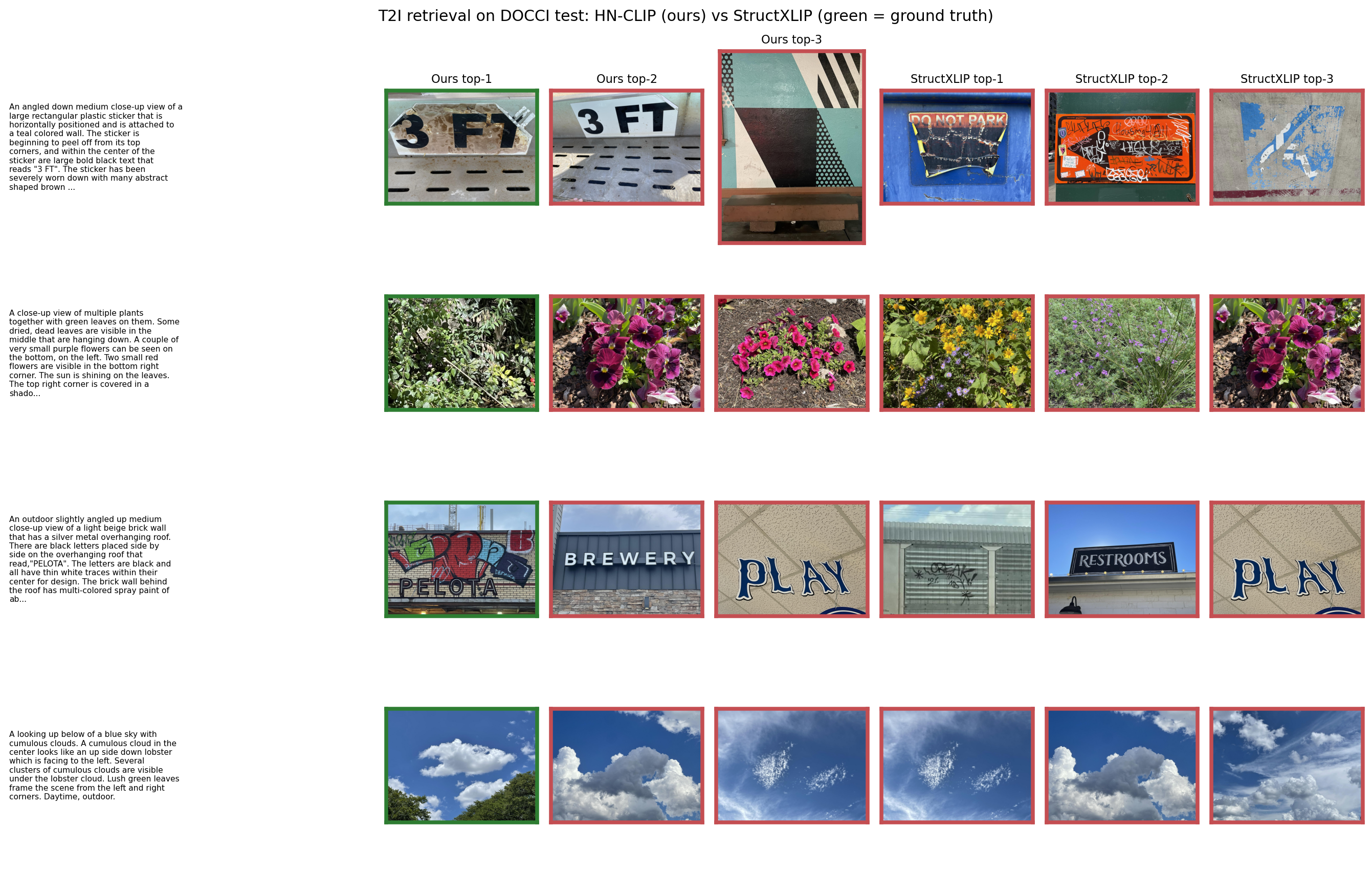}
  \caption{\textbf{Qualitative T$\rightarrow$I retrieval on DOCCI} (green
  = ground truth).}
  \label{fig:s-qual}
\end{figure}

\cref{fig:s-qual} shows Text$\rightarrow$Image retrievals on DOCCI
queries whose galleries are crowded with near-duplicates, the regime our
loss targets. \method{} resolves the discriminative details named in the
caption (lettering, flower species, signage), while the strongest
baseline retrieves appearance-level look-alikes, leaving the ground truth
at rank 18--52.

\end{document}